\documentclass[journal]{IEEEtran}
\usepackage{cite}
\usepackage{multicol}
\usepackage[hidelinks]{hyperref}
\usepackage{bm}
\usepackage{rotating}
\usepackage{url}
\usepackage{graphicx} 
\usepackage{mathptmx} 
\usepackage[mathscr]{euscript}  
\usepackage{amsmath} 
\usepackage{amssymb}  
\usepackage{multirow}
\usepackage{siunitx}
\usepackage{xcolor}
\usepackage{adjustbox}
\usepackage{array}
\usepackage{booktabs}
\usepackage[ruled,vlined]{algorithm2e}  

\usepackage[inline]{enumitem}
\usepackage[hang,flushmargin]{footmisc}
\usepackage{microtype}
\usepackage{threeparttable}
\usepackage[font=footnotesize,labelfont=bf]{caption}
\usepackage{pifont}
\usepackage{mathtools}

\DeclarePairedDelimiter\floor{\lfloor}{\rfloor}

\newcommand{\cmark}{\ding{51}}%
\newcommand{\xmark}{\ding{55}}%
\usepackage{verbatim}
\usepackage{xcolor}
\DeclareSIUnit{\degree}{deg}
\pdfoutput=1

\newcommand{\ls}[1]{\textit{Lov\'{a}sz-Softmax }{#1}}
\newcommand{\secref}[1]{Sec.~\ref{#1}}
\renewcommand{\eqref}[1]{Eq.~(\ref{#1})}
\newcommand{\figref}[1]{Fig.~\ref{#1}}
\newcommand{\tabref}[1]{Tab.~\ref{#1}}

\newcommand\myworries[1]{\textcolor{black}{#1}}

\newcolumntype{P}[1]{>{\centering\arraybackslash}p{#1}}

\begin{document}

\title{EfficientLPS: Efficient LiDAR Panoptic Segmentation}
\author{Kshitij Sirohi$^{*,1}$,
        Rohit Mohan$^{*,1}$,
        Daniel B\"uscher$^1$,
        Wolfram Burgard$^{1,2}$,
        and~Abhinav Valada$^1$
\thanks{$^*$These authors contributed equally.}%
\thanks{$^1$Department of Computer Science, University of Freiburg, Germany}
\thanks{$^2$Toyota Research Institute, Los Altos, USA.}}

%

\markboth{\copyright~2021 IEEE}
\IEEEaftertitletext{\vspace{-1\baselineskip}}


\maketitle

\begin{abstract}
Panoptic segmentation of point clouds is a crucial task that enables autonomous vehicles to comprehend their vicinity using their highly accurate and reliable LiDAR sensors. Existing top-down approaches tackle this problem by either combining independent task-specific networks or translating methods from the image domain ignoring the intricacies of LiDAR data and thus often resulting in sub-optimal performance. In this paper, we present the novel top-down Efficient LiDAR Panoptic Segmentation (EfficientLPS) architecture that addresses multiple challenges in segmenting LiDAR point clouds including distance-dependent sparsity, severe occlusions, large scale-variations, and re-projection errors. EfficientLPS comprises of a novel shared backbone that encodes with strengthened geometric transformation modeling capacity and aggregates semantically rich range-aware multi-scale features. It incorporates new scale-invariant semantic and instance segmentation heads along with the panoptic fusion module which is supervised by our proposed panoptic periphery loss function. Additionally, we formulate a regularized pseudo labeling framework to further improve the performance of EfficientLPS by training on unlabelled data. We benchmark our proposed model on two large-scale LiDAR datasets: nuScenes, for which we also provide ground truth annotations, and SemanticKITTI. Notably, EfficientLPS sets the new state-of-the-art on both these datasets.
\end{abstract}


%
\IEEEpeerreviewmaketitle

\section{Introduction}

\IEEEPARstart{A}{utonomous} vehicles are required to operate in challenging urban environments that consist of a wide variety of agents and objects, making comprehensive perception a critical task for robust and safe navigation. Typically, perception tasks are focused on independently reasoning about the semantics of the environment and recognition of object instances. Recently, panoptic segmentation~\cite{kirillov2019panoptic} which unifies semantic and instance segmentation has emerged as a popular scene understanding problem that aims to provide a holistic solution. Panoptic segmentation simultaneously segments the scene into 'stuff' classes that comprise of background objects or amorphous regions such as road, vegetation, and buildings, as well as 'thing' classes that represent distinct foreground objects such as cars, cyclists, and pedestrians. Panoptic segmentation has been extensively studied in the image domain~\cite{kirillov2019panoptic,mohan2020efficientps,porzi2019seamless,cheng2019panoptic}, facilitated by the ordered structure of images being supported by well-researched convolutional networks. However, only a handful of methods have been proposed for panoptic segmentation of LiDAR point clouds~\cite{gasperini2020panoster,milioto2020lidar}. LiDARs have become an indispensable sensor for autonomous vehicles due to their illumination independence and geometric description of the scene, making scene understanding using LiDAR point clouds an essential capability. However, the typical unordered, sparse, and irregular structure of point clouds pose several unique challenges.

To this end, deep learning methods that rely on grid based convolutions to address these challenges typically follow two different directions. They either project the point cloud into the 3D voxel space and employ 3D convolutions on them~\cite{maturana2015voxnet, graham20183d}, or they project the point cloud into the 2D space~\cite{cortinhal2020salsanext,milioto2020lidar,hurtado2020mopt} and employ the well-researched 2D Convolutional Neural Networks (CNNs). While voxel-based method achieve high accuracy, they are computationally more expensive and require substantial memory to store the voxelized point clouds. Methods such as~\cite{graham20183d,choy20194d} leverage the sparse nature of occupied voxel grids to improve the runtime and memory consumption. The 2D projection based methods on the other hand, yield a more denser representation and require comparatively lesser computational resources, but they suffer from information loss during projection, blurry CNN outputs, and incorrect label assignment to the occluded points during re-projection. Therefore, there is a need to bridge this gap with a method that has the advantages of fast and memory-efficient 2D convolutions while mitigating the problems due to the projection.

\begin{figure}
    \footnotesize
    \centering
    \includegraphics[width=0.47\textwidth]{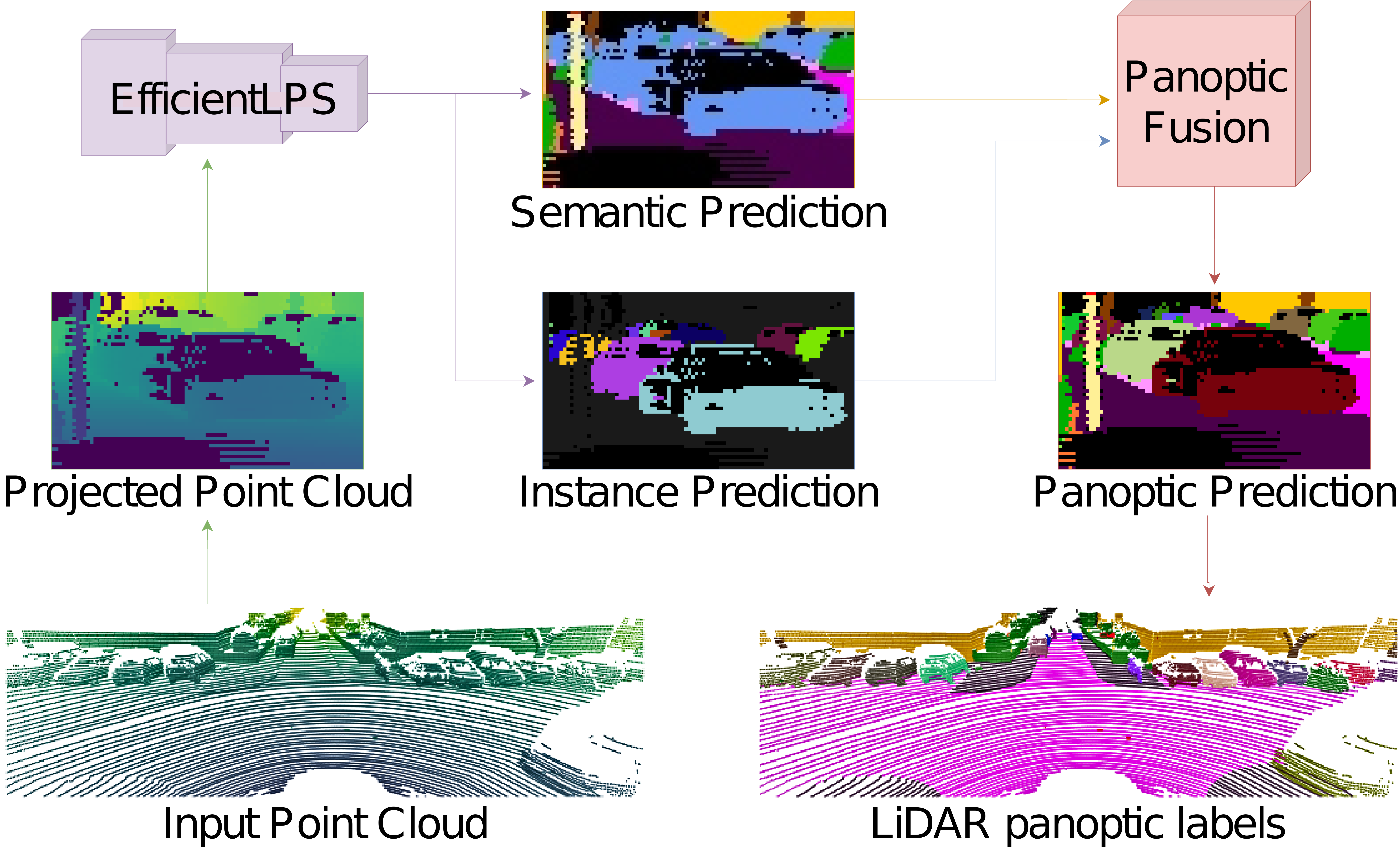}
    \caption{Overview of the top-down EfficientLPS architecture that consists of a shared backbone to learn spatially-aware features from the projected point cloud and individual heads to learn semantic and instance specific features which are fused in the panoptic fusion module. The network explicitly utilizes the range information in the backbone, semantic head and fusion module to mitigate the problems due to the projection and distance-dependent sparsity of LiDAR point clouds.}
    \label{fig:intropic}
\end{figure}

In this work, we present the novel Efficient LiDAR Panoptic Segmentation (EfficientLPS) architecture that effectively addresses the aforementioned challenges by employing a 2D CNN for the task while explicitly utilizing the unique 3D information provided by point clouds. EfficientLPS consists of a shared backbone comprising our novel Proximity Convolution Module (PCM), an encoder, the proposed Range-aware FPN (RFPN) and the Range Encoder Network (REN). We build the encoder and REN based on the EfficientNet~\cite{tan2019efficientnet} family, therefore we follow the convention of naming our model with the Efficient prefix. EfficientLPS also consists of a novel distance-dependent semantic segmentation head and an instance segmentation head, followed by a fusion module that provides the panoptic segmentation output. Our network makes several new contributions to address the problems that persist in LiDAR cylindrical projections. We propose the Proximity Convolution Module (PCM) that alleviates the problems caused by the fixed geometric grid structure of a standard convolution which is incapable of modeling object transformations such as scaling, rotation and deformation~\cite{dai2017deformable}. The problem of distance-dependent sparsity further exacerbates the limited transformation modeling capability of standard convolutions. The PCM models the transformations of objects in the scene by leveraging the contributions of nearby points, effectively reshaping the convolution kernel depending on range values.

When LiDAR points are projected into the 2D domain, objects tend to be closer to each other. Hence, the network in these cases often ignores smaller objects in favor of larger overlapping objects. Although this overlap is more distinguishable in the range channel of the projections, the features computed over all the projection channels begin to lose track of this distinction as they try to capture more and more complex representations in the deeper layers of the encoder. To alleviate this problem and enable the network to better distinguish adjacent objects, we propose the Range-aware Feature Pyramid Network (RFPN). We employ the REN parallel to the encoder to solely encode the range channel of the projection and selectively fuse it with the FPN outputs to compute range-aware multi-scale features.

Moreover, there is a large variation in the scale of objects due to the projection of the point cloud into the 2D domain. The objects that are closer tend to be larger in scale, and objects at a farther distance tend to be smaller. Hence, the 2D projection consists of objects that have distance-dependent scale variations. Typically, the instance head of top-down methods cope with it to a certain extent using many predefined anchors at different scales. However, the semantic head that predominantly aggregates multi-scale features tends to suffer~\cite{chen2018encoder,chen2018searching, mohan2020efficientps}. In order to mitigate this problem, we propose the distance-dependent semantic head that consists of modules that incorporate our range-guided depth-wise atrous separable convolutions in addition to fixed multi-dilation rate convolutions to generate features that cover a relatively large scale range in terms of the receptive field in a dense manner.

Furthermore, segmented objects in the projection domain often tend to have inaccurate boundaries. In the image domain, these inaccuracies only span a few pixels so they have little or negligible effect on the overall performance. However, when the segmented output in the projection domain is re-projected back into point clouds, it causes leakage of object boundaries into the background and significantly affects the performance of the model. To address this problem, we introduce the novel panoptic periphery loss function that operates on the logits of the panoptic fusion module to effectively combine the outputs of both heads. Our proposed loss function refines 'thing' instance boundaries by maximizing the range separation between the foreground boundary pixels, i.e., the 'thing' instance boundary and the neighboring background pixels.

Most supervised learning methods require large amounts of annotated training data and manually labeling point clouds is an extremely arduous task. As an alternative solution to this problem, we explore the viability of generating pseudo labels from the abundantly available unlabeled point cloud datasets. We formulate a new framework for computing regularized pseudo labels from unlabeled data, given some labeled data with similar properties. The regularized pseudo labels aim to reduce the incorrect predictions on the unlabeled dataset to prevent confirmation bias. To the best of our knowledge, this is the first work to propose a pseudo labeling technique for any point cloud scene understanding task.

We perform extensive evaluations on the SemanticKITTI~\cite{behley2019iccv} and nuScenes~\cite{caesar2020nuScenes} datasets which have point clouds with different densities to demonstrate the generalization ability of our model. As the nuScenes dataset itself does not provide panoptic segmentation labels, we compute the annotations from the publicly available semantic segmentation and 3D bounding box annotations. We provide several baselines and make the nuScenes panoptic segmentation dataset publicly available to encourage future research using sparse point clouds. Our proposed EfficientLPS consistently outperforms existing methods, thereby setting the new state of the art on both datasets and is ranked \#1 on the SemanticKITTI leaderboard. Finally, we present detailed ablation studies that demonstrate the novelty of the various architectural contributions that we make in this work.

To summarize, the main contributions of this work are as follows:
\begin{enumerate}
  \item A novel top-down architecture that consists of a shared backbone with task-specific heads that incorporate our proposed range enforced components and a fusion module supervised by our panoptic periphery loss function.
  \item The proximity convolution module which boosts the transformation modeling capacity of the shared backbone by leveraging range proximity between neighboring points. 
  \item The novel range-aware feature pyramid network that reinforces bidirectionally aggregated semantically rich multi-scale features with spatial awareness.
  \item The new semantic head that captures scale-invariant rich characteristic and contextual features using our range-guided depth-wise atrous separable convolutions.
  \item The novel panoptic periphery loss function that refines the segmentation of 'thing' instances by maximizing the range separation between foreground boundary pixels and neighboring background pixels.
  \item A new framework for improving panoptic segmentation of LiDAR point clouds by exploiting large unlabelled datasets via regularized pseudo label generation.
  \item Exhaustive quantitative and qualitative evaluations of our model along with comprehensive ablation studies of our proposed architectural components.
  \item We made the code and models publicly available at \url{http://rl.uni-freiburg.de/research/lidar-panoptic}
\end{enumerate}

\section{Related Works}
\label{sec:realtedwork}

Panoptic segmentation has emerged as a popular scene understanding task since its introduction by Kirillov~\textit{et~al.}~\cite{kirillov2019panoptic}. By unifying semantic segmentation and instance segmentation, it aims at holistic scene understanding and reasoning. While panoptic segmentation has been extensively studied in the 2D domain using RGB images, only a handful of methods address this task in the 3D domain of LiDAR point clouds. In the following, we first discuss different convolution kernels and different techniques of representing point cloud data. We then discuss recent works that address the various scene segmentation tasks using point clouds in autonomous driving scenarios, namely: semantic, instance, and panoptic segmentation.


\subsection{Convolution kernels}

Standard 2D convolutions lack the ability to model geometric relations in the 3D domain due to the nature of the convolution operation that samples from fixed locations. Some works model these relations with the help of non-regular grids~\cite{dai2017deformable} where the learned offsets change the shape of the sampling grid in convolutions. Using RGB-D images, Wang~\textit{et~al.}~\cite{wang2018depth} directly use the depth value to weight the contribution of neighboring pixels for the convolution output. While the convolution shaping methods using RGB images learn the lacking spatial information, we propose a shaping mechanism using the already available spatial information in point clouds. We devise the Proximity-aware Convolution Module (PCM) that reshapes the convolution grid to capture local contextual information from neighboring pixels. This is especially helpful for capturing contextual information of very distant points in LiDAR point cloud that suffer from distance-induced sparsity. 

\subsection{Data Representation}


The methods that are typically employed on point clouds can be broadly classified into three categories, namely: point-based, volumetric, and projection-based techniques. PointNet~\cite{qi2017pointnet} is one of the first pioneering point-based methods which learns features using MLPs followed by max-pooling to extract global context.
More recent methods~\cite{boulch2020convpoint, xu2018spidercnn, thomas2019kpconv} develop convolution operations and kernels specially designed to work with point clouds. Kernel Point Convolution (KPConv)~\cite{thomas2019kpconv} is one such method with flexible kernel points in the 3D space, which are learned in a similar manner on point clouds as in 2D convolutions. On the other hand, volumetric methods transform the point clouds into regular voxel grids and apply 3D convolutions~\cite{maturana2015voxnet}. Most methods that rely on 3D convolution are both memory and computationally intensive which limits the resolution of the voxels and hence the overall performance. However, to account for the sparse nature of the voxel grid, sparse 3D convolutions~\cite{graham20183d,choy20194d} have been proposed to decrease the runtime and memory footprint.

In projection-based methods, the point cloud is projected onto an intermediate regular 2D grid representation to facilitate the use of well researched 2D convolution architectures. To obtain pseudo sensor data such as the grid representation, existing methods project points either using the spherical projection~\cite{milioto2019rangenet++,wu2018squeezeseg} or using scan unfolding~\cite{triess2020}. Conversely, point clouds are also projected into a bird's eye view (BEV)~\cite{zhang2020polarnet} to exploit the radial nature and obtain better spatial segregation. In this work, we employ scan unfolding due to its ability to recover dense representations, similar to the original format that a LiDAR sensor provides.

\subsection{Scene Understanding using LiDAR Point Clouds}

\subsubsection{Semantic Segmentation}

The challenges posed by the unordered and sparse nature of point clouds has hindered the progress in LiDAR semantic segmentation for autonomous driving. Considerably lesser number of techniques have been proposed to address this task using point clouds in comparison to methods in the visual domain.
Dewan~\textit{et~al.}~\cite{dewan2017deep} propose an approach to classify points into movable, non-movable, and dynamic classes, by combining deep learning-based semantic cues and rigid motion based motion cues in a Bayesian framework. Wu~\textit{et~al.}~\cite{wu2018squeezeseg} propose a projection-based approach that builds upon SqueezeNet and introduces the fire module which is incorporated into the encoder and decoder. DeepTemporalSeg~\cite{dewan2020deeptemporalseg} employs Bayesian filtering to obtain temporally consistent semantic segmentation.

The release of the SemanticKITTI~\cite{behley2019iccv} dataset motivated many works in semantic segmentation of LiDAR point clouds. Milioto~\textit{et~al.}~\cite{milioto2019rangenet++} propose a 2D CNN architecture that operates on spherically projected point clouds and employs a kNN based post-processing step to account for the occlusions due to the projection. SalsaNext~\cite{cortinhal2020salsanext} uses spherical projection for semantic segmentation and also performs uncertainty estimation. PolarNet~\cite{zhang2020polarnet} projects the point cloud in the birds-eye view and employs a ring convolutions on the radially defined grids. Some methods follow a hybrid approach by combining 2D and 3D convolutions. KPRNet~\cite{kochanov2020kprnet} uses 2D convolutions for semantic segmentation followed by KPConv-based~\cite{thomas2019kpconv} post-processing using point-wise convolutions. SPVNAS~\cite{tang2020searching} automates architecture design by employing neural architectural search to search for efficient 3D convolution-based models.

\subsubsection{Instance Segmentation}

Similar to the image domain, instance segmentation of point clouds can be classified into two categories: proposal based and proposal free methods. Proposal based methods perform 3D bounding box detection followed by point-wise mask generation for the points in each bounding box. 3D~Bonet~\cite{yang2019learning} follows this approach using two separate 3D bounding box proposal generation and mask generation branches. GSPN~\cite{yi2019gspn} generates proposals using a shape aware proposal generation for different instances. On the other hand, proposal free methods directly predict the instances by detecting keypoints such as the centriod of the instance, or the similarity between points, which is followed by clustering~\cite{zhang2020instance}. SGPN~\cite{wang2018sgpn} learns a similarity matrix between the points which is used to cluster points with higher similarity scores between them. PointGroup~\cite{jiang2020pointgroup} extracts semantic information using 3D sparse convolutions to cluster points towards the instance centroid, followed by using the original points and the clustered points to obtain the final prediction. VoteNet~\cite{ding2019votenet} predicts an offset vector to the centroid of every point and then employs clustering.

\begin{figure*}
\centering
\includegraphics[width=1.0\textwidth]{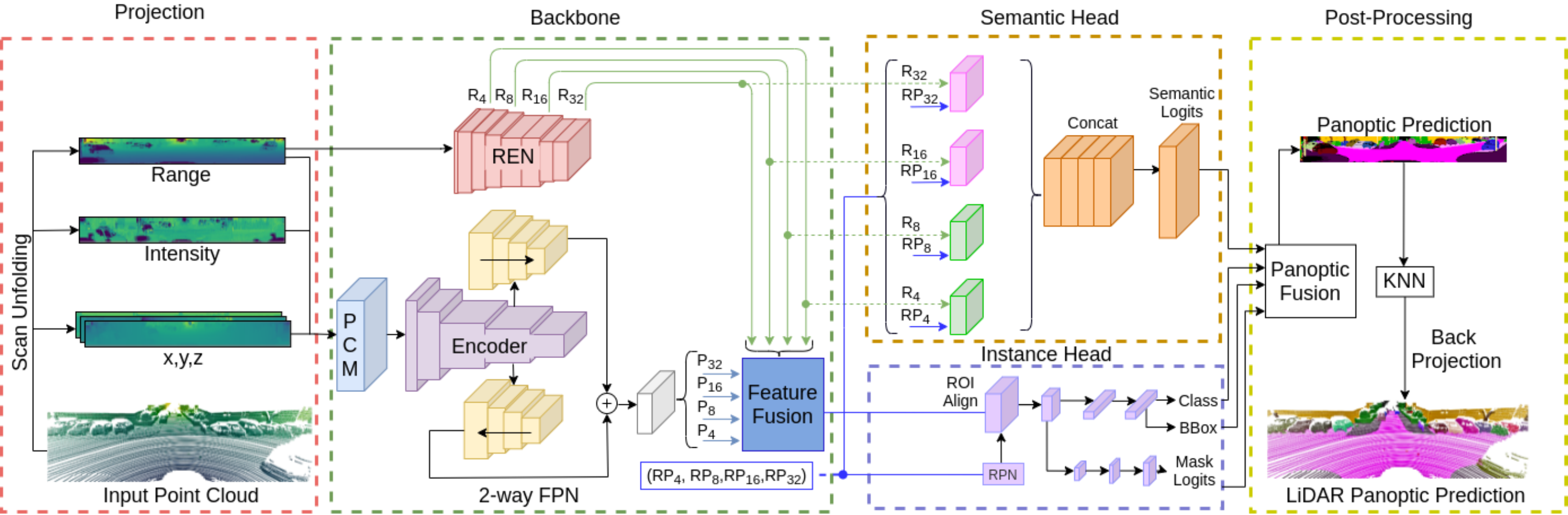}
\caption{Illustration of our proposed EfficientLPS architecture for LiDAR panoptic segmentation. The point cloud is first projected into the 2D domain using scan unfolding and fed as an input to our Proximity Convolution Module (PCM). Subsequently, we employ the shared backbone consisting of the EfficientNet encoder with the 2-way FPN and the Range Encoder Network (REN) in parallel. The output of these two modules are fused and fed as input to the semantic and instance heads. The logits from both heads are then combined in the panoptic fusion module which is supervised by the panoptic periphery loss function. Finally, the output of the panoptic fusion module is projected back to the 3D domain using a kNN algorithm.}
\label{fig:network}
\end{figure*}

\subsubsection{Panoptic Segmentation}

Panoptic segmentation methods can also be classified into proposal-free (bottom-up) or proposal-based (top-down) techniques. Bottom-up methods group points belonging to the same instances either by a voting scheme or based on pixel-pair affinity pyramid while simultaneously learning the semantic labels~\cite{gao2019ssap}. On the other hand, top-down approaches~\cite{hurtado2020mopt} tackle the problem in two a stage manner with a dedicated instance segmentation branch for detecting and segmenting~'thing' classes, and a semantic segmentation branch for segmenting the~'stuff' classes.

Miliotto~\textit{et~al.}~\cite{milioto2020lidar} and Gasperini~\textit{et~al.}~\cite{gasperini2020panoster} adopt the bottom-up approach where instances are detected without region proposals. Miliotto~\textit{et~al.} use spherical projection of point clouds and predict offsets to the centroids for aiding clustering. They also use 3D information available in the range images for trilinear upsampling in the decoder. Panoster~\cite{gasperini2020panoster} uses an instance head which directly provides the instance ids of the points from learnable clustering without any explicit grouping requirement. In addition to the spherical projection-based method Rangenet++~\cite{milioto2019rangenet++} and Panoster~\cite{gasperini2020panoster} also show the implementation of their clustering mechanism using the point-based method KPConv~\cite{thomas2019kpconv} for semantic segmentation. PanopticTrackNet~\cite{hurtado2020mopt} further unifies panoptic segmentation with multi-object tracking and provides temporally consistent instance labels.

\subsubsection{Semi-Supervised Learning}

Semi-supervised learning (SSL) for LiDAR panoptic segmentation has not been explored. Thus, we discuss the two prominent SSL approaches for 3D object detection. SESS~\cite{zhao2020sess} trains a EMA-based teacher model and a student model simultaneously with consistency loss between them whereas 3DIoUMatch~\cite{wang20213dioumatch} employs mean-teacher~\cite{tarvainen2017mean} based framework using an IoU prediction filtering mechanism. Both of the approaches use augmentation of point clouds for which the point cloud is available. In contrast, our proposed pseudo labeling framework exploits external unlabeled point cloud datasets with a separate teacher and student model.

In this work, we present a novel LiDAR panoptic segmentation network that effectively exploits the advantages of projection-based top-down methods. Our proposed architecture comprises of a shared backbone that incorporates the proposed proximity convolution module in the beginning to boost its geometric transformation modeling capacity and the novel range-aware FPN at the end to capture spatially aware and semantically rich multi-scale features. It further consists of a modified Mask~R-CNN~\cite{he2017mask} instance head and a new semantic head that fuses distance-dependent fine and long-range contextual features with distance-independent features for enhanced scale-invariance. We also propose the novel panoptic periphery loss function that refines the 'thing' object class boundaries by maximizing the range difference between the foreground and background pixels of the instance boundaries. All of the aforementioned modules effectively leverage the intricacies of LiDAR data to address the issues prevalent in projection-based LiDAR segmentation. Additionally, we explore the viability of pseudo labeling for LiDAR panoptic segmentation and thus propose a novel regularized pseudo labeling framework for the same.

\section{Technical Approach}
\label{sec:technical}

In this section, we present a brief overview of our proposed EfficientLPS architecture and then detail each of its constituting components. \figref{fig:network} illustrates the topology of EfficientLPS that follows the top-down layout. First, we project the point cloud from the LiDAR scanner into the two-dimensional space using scan unfolding~\cite{triess2020}. The projected representation comprises of five channels: range, intensity and the $(x,y,z)$ coordinates. We then employ our novel shared backbone which consists of our proposed Proximity Convolution Module (PCM) to aid in modeling geometric transformations, followed by a modified EfficientNet-B5 encoder with the 2-way~FPN~\cite{mohan2020efficientps}. We employ the proposed Range Encoder Network (REN) in parallel which takes the range channel of the projected point cloud as input. We then fuse the multi-scale outputs of the encoder and REN to obtain the range-aware feature pyramid that enhances the ability to distinguish adjacent objects at different distances. The entire shared backbone is enclosed with green dashed lines in \figref{fig:network}.\looseness=-1

Following the backbone, we employ the parallel semantic segmentation (depicted in orange) and instance segmentation (depicted in purple) heads. The semantic head consists of Dense Prediction Cells (DPC)~\cite{chen2018searching} and Large Scale Feature Extractor (LSFE)~\cite{mohan2020efficientps} units, which we extend with our proposed range-guided depth-wise atrous separable convolutions to enable capturing scale invariant long-range contextual and fine features. We use a variant of Mask~R-CNN~\cite{he2017mask} for the instance head. We then fuse the logits from both the heads in the panoptic fusion module~\cite{mohan2020efficientps} which is supervised by our panoptic periphery loss to facilitate object boundary refinement and yield the panoptic segmentation in the projection domain. Finally, we re-project the predictions into the 3D space to obtain the final panoptic segmentation output of the input point cloud. During training, we employ our proposed pseudo labeling technique to train our model with both labeled and unlabeled data. In the rest of this section, we describe each of the aforementioned components in detail.


\subsection{Projection using Scan Unfolding}
\label{sec:scanUnfold}

We employ scan unfolding~\cite{triess2020} to project the point cloud into the 2D range image format. Scan unfolding aims to mitigate the problems due to the alternate spherical projection method which suffers from point occlusions due to ego-motion correction. Scan unfolding yields a much denser representation than spherical projection. LiDAR sensors typically provide raw data in a range image-like format, with each pixel describing the range value at a particular row and column. Each column of this range image consists of measurements taken by individual modules stacked vertically within the sensor at a particular time and each row represents the consecutive measurements of one module taken during spinning of the sensor. However, most of the publicly available datasets provide LiDAR measurements as a list of 3D Cartesian coordinates, without any information about the column or row indices. Hence, we assign these indices to every point in the laser scan for recovering the range image-like representation.

In order to project the LiDAR scan represented as a point cloud to the range image, we assign row and column indices to every point in the scan corresponding to an image of size $W \times H$. The list of points provided by datasets such as KITTI~\cite{Geiger2013IJRR} is typically constructed by just concatenating the rows of each scan. Hence, it is still ordered by horizontal and vertical indices. The scan unfolding algorithm takes advantage of this information and sequentially computes the yaw difference between the consecutive points to recover the vertical index. A jump is detected when the yaw difference is above a pre-defined threshold, which increments the vertical index of the following points in the list.
We chose a threshold value of $\SI{310}{\degree}$, since the yaw angle typically drops from a value near $\SI{360}{\degree}$ to near $0$ between the rows.
The horizontal index is computed as $\floor{(0.5 (1-\phi/\pi) W)}$, where $\phi$ is the yaw angle of each point in the range $[-\pi, \pi]$.
The points are projected to their corresponding rows and columns with range, intensity and $(x,y,z)$ coordinates represented as separate channels. This results in a tensor of shape $(5 \times H \times W)$ which is fed as input to the network. Note that the value of $H$ is the number of vertically placed sensor modules in a sensor, which is 64 for the KITTI dataset. Other datasets that already contain the vertical index information, such as nuScenes, only require the computation of the horizontal index. 

\subsection{EfficientLPS Architecture}
\subsubsection{Backbone}

The backbone consists of our proposed Proximity Convolution Module (PCM), followed by an encoder and the novel Range-aware Feature Pyramid Network (RFPN). We detail each of these components in the following section.

\noindent\textbf{Proximity Convolution Module:} The core of the PCM is the proximity convolution operation. The standard convolution operation performs sampling over a feature map followed by a weighted sum of the sampled values to yield an output feature map $y$. The convolution at pixel $p$ is computed as
\begin{equation}
y(p)=\sum_{p_o\in{R}}w(p_o)\cdot x(p+p_o),
\label{eq:standard_conv}
\end{equation}
where ${R}$ is a regular sampling grid containing the sampling offset locations around $p$ in the input feature map $x$ weighted by the kernel $w$.

The standard convolution is limited in its geometric transformation modeling capacity due to its fixed grid structure. The distance-dependent sparsity present in the LiDAR data further exacerbates the effects of this limitation. To tackle this constraint, we propose the proximity convolution which exploits range information to augment the spatial sampling locations for effectively improving the transformation modeling ability. Formally, in the proximity convolution, for each pixel $p$ in the projected range image $R \in \mathbb{R}^{h \times w}$, we compute its nearest neighbors using the k-Nearest-Neighbors (kNN) algorithm. Here, we use range difference of the corresponding points from range image as the distance in the 3D space to find the nearest neighbors. Subsequently, we sort the nearest neighbors in the ascending order of their range difference to the query pixel. We now adapt \eqref{eq:standard_conv} as
\begin{equation}
y(p)=\sum_{p_n\in{N}}w(p_n)\cdot x(p+p_n),
\label{eq:range_conv}
\end{equation}
where ${N}$ is no longer a regular grid but consists of offsets for the \textit{n} nearest neighbors of pixel $p$. Please note that the grid ${N}$ includes the offset to the query pixel and its $n-1$ nearest neighbors. The weights $w$ are learned in the same manner as in standard convolutions. The search grid for the kNN algorithm is always larger than the learnable weight matrix and value of $k$ is the product of kernel size dimensions.

\begin{figure}
\centering
\includegraphics[width=0.45\textwidth]{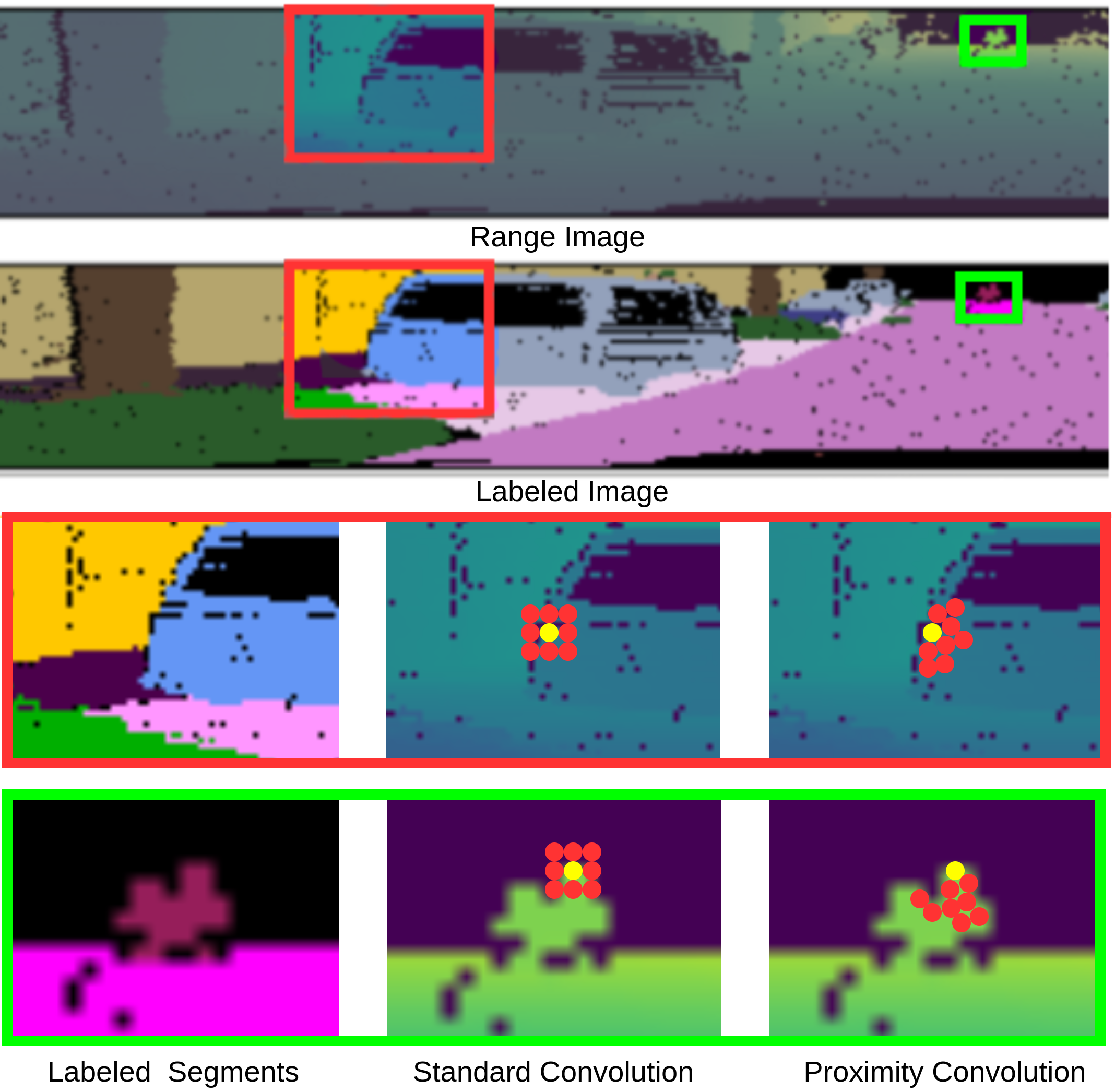}
\caption{Comparison of pixel sampling while applying standard convolution and our proposed proximity convolution. The highlighted red box contains a car and green box contains a bike. The convolution is applied by placing the kernel at the yellow dot, and the sampled neighboring pixels are represented by red dots. Observe that the neighboring pixels are sampled adaptively based on the range difference of the corresponding points in the range image.} 
\label{fig:PAC}
\end{figure}

The sampling operation of the proximity convolution in comparison to the standard convolution is illustrated in \figref{fig:PAC}. In the example, the convolutions are performed at the border of the objects. As shown in \figref{fig:PAC}, the proximity convolution forms the kernel according to the shape of the object, while the standard convolution obtains information outside the objects. Particularly, the standard convolution is not able to obtain information for the bike rider (green rectangle), since the large distance of the object from the sensor causes increased sparsity. Since the proximity convolution is less dependent on this distance-induced sparsity, it successfully represents the shape of the bike rider. Therefore, the proximity convolution models the geometric transformations more effectively, especially for farther away objects that suffer from distance-induced sparsity.

The proximity convolution module comprises of the proposed proximity convolution, synchronized Inplace Activated Batch Normalization (iABNsync)~\cite{porzi2019seamless} and Leaky ReLU activation. We use iABNsync in this layer and all the subsequent parts of the network in contrast to the vanilla batch normalization layer, as it provides a better estimate of the gradients and reduces the GPU memory footprint. We study the performance of the proximity convolution module in the ablation study presented in \secref{sec:pil}. 

\noindent\textbf{Encoder:} We adopt the EfficientNet~\cite{tan2019efficientnet} topology for the main encoder as well as the Range Encoder Network (REN). We remove the Squeeze and Excitation~(SE) connections to enable better localization of features and contextual elements. Similar to the proposed proximity convolution module, we replace the batch normalization layers with iABNsync and Leaky ReLU activation. The EfficientNet architecture comprises of nine blocks where blocks 2, 3, 5, and 9 yield multi-scale features that correspond to the down-sampling factors of $\times4$, $\times8$, $\times16$ and $\times32$ respectively.

EfficientNet employs compound scaling to scale the base network efficiently. Here, width, depth, and the resolution of the network are the coefficients available for scaling. We choose the scaling coefficients for the main encoder as 1.6, 2.2, and 456 respectively and the coefficients for the REN as 0.1, 0.1, and 224 respectively, which we obtain via grid search optimization. The output of the PCM is fed as input to the main encoder and the REN takes the projected range image as input. \figref{fig:network} depicts the main encoder in light purple and the REN in red.

\begin{figure*}
    \centering
    \includegraphics[width=1\textwidth]{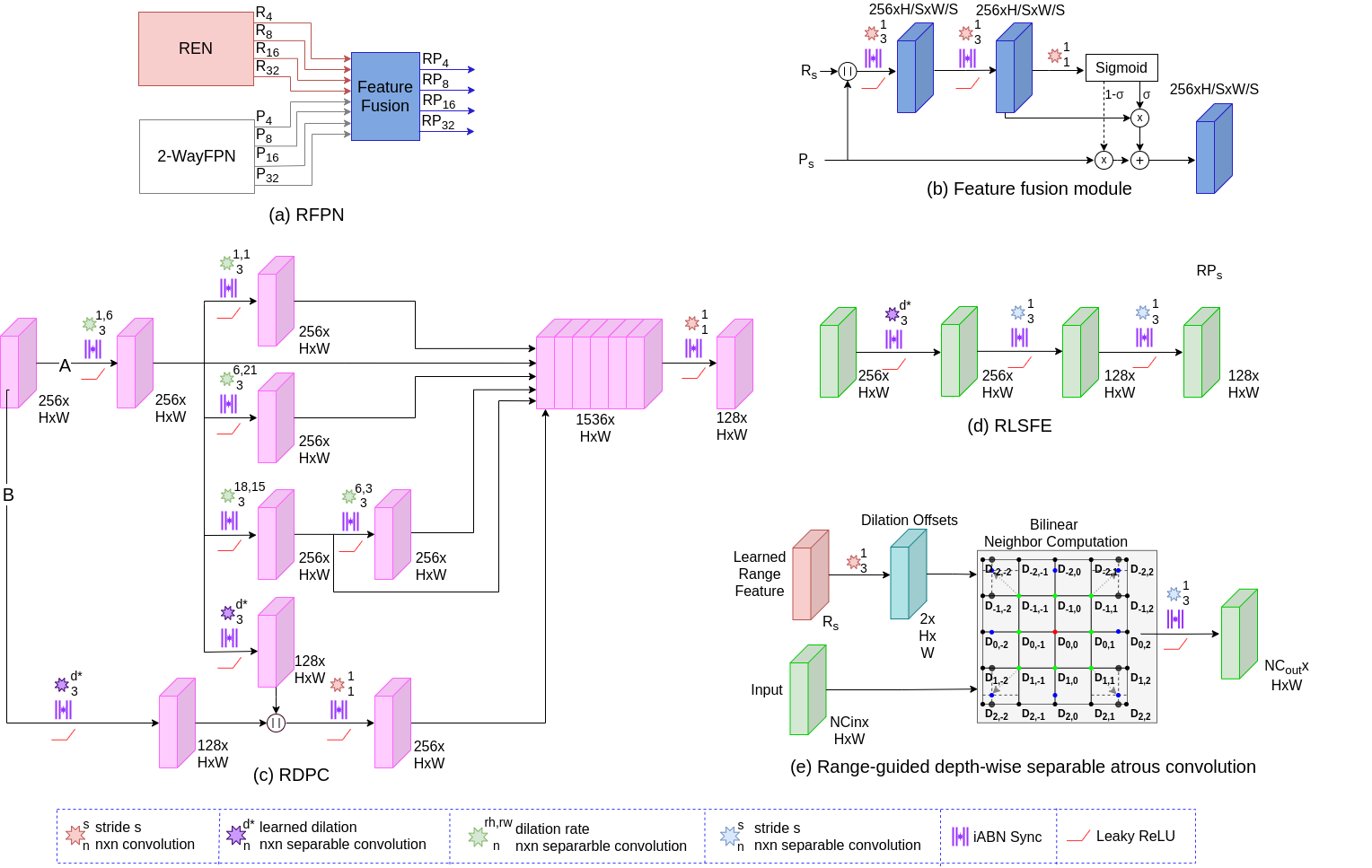}
    \caption{Topology of the different proposed architectural components in EfficientLPS. (a)~Range-Aware FPN (RFPN) and (b) Feature fusion module is used for the fusion of range encoded features with FPN features in RFPN. (c) Range-guided Dense Prediction Cells (RDPC) and (d) Range-guided Large Scale Feature Extractor (RLSFE) modules are part of the proposed semantic head. Lastly, (e) Range-guided depth-wise atrous separable convolution is the mechanism for controlling dilation offsets in the RDPC and RLSFE modules.}
    \label{fig:components}
\end{figure*}

\noindent\textbf{Range-Aware FPN:} Our proposed range-aware FPN (RFPN) reinforces the coherently aggregated fine and contextual features of Feature Pyramid Networks (FPNs) with distance awareness. This enables the network to better segregate adjacent objects with different range variations. We build upon the 2-way FPN~\cite{mohan2020efficientps} that enables bidirectional flow of information using two parallel branches that aggregate multi-scale from the main encoder in a top-down and bottom-up manner respectively. The 2-way FPN is depicted with yellow blocks in \figref{fig:network}. The outputs from both the parallel branches at each resolution are summed together and passed through a $3\times3$ convolution with 256 output channels to yield the outputs: P\textsubscript{4}, P\textsubscript{8}, P\textsubscript{16}, and P\textsubscript{32}. Note that we use standard convolutions in the 2-way FPN instead of separable convolutions used in~\cite{mohan2020efficientps} to learn richer representations at the expense of additional parameters.

Our proposed RFPN consists of the aforementioned 2-way FPN, the REN module, and the feature fusion module as shown in \figref{fig:components}~(a). The REN module is employed in parallel to the 2-way FPN. We find that enabling the REN to learn to encode range information explicitly in different scales rather than direct downsampling of range data, yields better performance as shown in the ablation studies presented in \secref{sec:RFPN}. The outputs of REN which are at four different resolutions (R\textsubscript{4}, R\textsubscript{8}, R\textsubscript{16} and R\textsubscript{32}) and the outputs of the 2-way FPN (P\textsubscript{4}, P\textsubscript{8}, P\textsubscript{16} and P\textsubscript{32}) are fed as input to the Feature Fusion module which computes range-aware pyramid features for each of the corresponding resolution (RP\textsubscript{4}, RP\textsubscript{8}, RP\textsubscript{16} and RP\textsubscript{32}). Here, 4, 8, 16 and 32 denote different downsampling factors with respect to the input.  

The purpose of the fusion module is two-fold. First, to fuse the inputs R\textsubscript{s} and P\textsubscript{s}, where s denotes the downsampling factor. Second, to enable the network to emphasize on the more informative features between the 2-way FPN features and the corresponding fused features, hence incorporating distance awareness selectively. As shown in \figref{fig:components}~(b), the feature fusion module consists of two branches. The first branch takes the concatenated tensors R\textsubscript{s} and P\textsubscript{s} as input, and feeds them through two $3\times3$ convolution layers sequentially to yield the fused features G\textsubscript{s}. Additionally, we use iABNsync and leaky ReLU layers after each $3\times3$ convolution. We compute the weight factors for this branch ($w_{fs}$) by employing a $1\times1$ convolution followed by a sigmoid activation. The second branch propagates P\textsubscript{s} in parallel and the weight factor of this branch is computed as $1-w_{fs}$. Then the output (RP\textsubscript{s}) of this module is given by
\begin{equation}
{RP}_{s} = w_{fs}*G_{s}+(1-w_{fs})*P_{s},
\label{eq:range_fpn}
\end{equation}

For the sake of simplicity, the fusion module is depicted as one blue box in \figref{fig:network}, whereas in practice each resolution has its own exclusive feature fusion module depicted in \figref{fig:components}~(b). We present detailed analysis of different components of the range-aware~FPN in the ablation study in \secref{sec:RFPN}.

\subsubsection{Distance-Dependent Semantic Head}

The main component of our proposed distance-dependent semantic head is the novel range-guided depth-wise atrous separable convolution operation. We essentially encode the range using the REN module and compute the dilation factor to apply at each central pixel from the encoded features. We then employ the depth-wise atrous separable convolution operation with the computed dilation factor, thereby enabling the receptive field to be adaptable to the range data. As shown in \figref{fig:components}~(e), we employ a $3\times3$ convolution on the encoded range features from the REN module to obtain the dilation offsets for each pixel. Subsequently, we compute the bilinearly interpolated neighbors from the input at the corresponding offsets for each pixel. We then employ a $3\times3$ depth-wise separable convolution on the computed neighbors to generate the final output. Note that the input, the REN encoded features, and the dilation offsets have the same spatial resolution. We also use a parameter $D_{max}$ in this convolution to set the maximum value for dilation offsets.

The range-guided depth-wise atrous separable convolution learns scale-invariant features as the dilation rate of the convolution kernel changes based on the distance, and so does the scale of objects. We take advantage of this scale invariance in the proposed semantic head of our EfficientLPS architecture. We extend the semantic head proposed in~\cite{mohan2020efficientps} consisting of Dense Prediction Cells (DPC), Large Scale Feature Extractor (LSFE), and Mismatch Correction Module (MC) with bottom-up path augmentation connections. We effectively retain the MC module and the bottom-up path augmentation connections but redesign the DPC and LSFE modules by incorporating our range-guided depth-wise atrous separable convolutions. We refer to this new LSFE variant as Range-guided Large Scale Feature Extractor (RLSFE) that comprises a $3\times3$ range-guided depth-wise atrous separable convolution with 256 output filters and $D_{max}=3$ followed by an iABNsync and a Leaky ReLU activation function. The $D_{max}$ parameter is set to a lower value in this module as it captures fine features that can get distorted with higher dilation rates. We employ two $3\times3$ separable convolutions with 128 output filters, each followed by iABNsync and a Leaky ReLU activation to further perform channel reduction and learn deeper features. \figref{fig:components}~(d) shows the topology of our proposed RLSFE module.

The topology of our Range-guided Dense Prediction Cells (RDPC) module is depicted in \figref{fig:components}~(c). We refer to the $3\times3$ depth-wise separable convolution with 256 output channels and a dilation rate of (1,6) as branch A. Similarly, the parallel $3\times3$ range-guided depth-wise atrous separable convolution with 128 output filters with $D_{max}=24$ is referred to as branch B. Branch A further splits into four parallel branches. Three of the branches consist of a $3\times3$ depth-wise separable convolution with 256 output channels with dilation rates (1,1), (6,21), and (18,15) respectively. The fourth branch is an identity connection that goes to the end in parallel to all the branches. The fifth branch concatenates with Branch B and consists of a $3\times3$ range-guided depth-wise atrous separable convolution with 128 output filters with $D_{max}=24$, and runs parallel to other branches. The branch with (18,15) dilation rates further branches out into an identity connection and a branch with a $3\times3$ depth-wise separable convolution with 256 output channels and dilation rate (6,3). In the end, there are a total of six parallel branches that are concatenated together to yield a tensor with 1536 channels. Finally, a $1\times1$ convolution with 256 output channels generates the output of the RDPC module. Please note that each of the aforementioned convolutions in the RDPC module is followed by an iABNsync and Leaky ReLU activation.  
 
The RDPC module essentially integrates range-guided depth-wise atrous separable convolutions with fixed multi-dilation rates ones to generate features that cover a relatively large scale range in terms of receptive field in a dense manner. In summary, our proposed distance-dependent semantic head consists of two RDPC modules employed at $\times32$ and $\times16$ downsampling factor whose inputs are $RP_{32}$, $R_{32}$ and $RP_{16}$, $R_{16}$ respectively. It also utilizes two RLSFE modules for $\times8$ and $\times4$ downsampling factor with $RP_{8}$, $R_{8}$ and $RP_{4}$, $R_{4}$ as inputs. These modules are subsequently followed by the MC module~\cite{mohan2020efficientps} with bottom-up path augmentations. In the last step, we apply a $1\times1$ convolution with $N_{stuff+thing}$ output filters and upsample to the resolution of the input image.
We train our semantic head with equally weighted per-pixel log-loss (${L}_{pp}$)~\cite{bulo2017loss} and \textit{Lov\'{a}sz-Softmax}~loss (${L}_{LS}$)~\cite{berman2018lovasz} as
\begin{equation}\label{eq:Lsemantic}
{L}_{semantic}(\Theta)= {L}_{pp} + {L}_{LS},
\end{equation}
We evaluate the performance of our proposed semantic head in the ablation study presented in \secref{sec:semantichead}.

\subsubsection{Instance Head}

We adopt the Mask~R-CNN~\cite{he2017mask} topology for the instance head and make certain modifications. We replace the batch normalization and ReLU activations with iABNsync and Leaky ReLU layers respectively. \figref{fig:network} shows the instance head depicted in purple blocks which consists of two stages. In the first stage, the Region Proposal Network (RPN) employs a fully convolutional network to generate object proposals and objectness scores for each output resolution of the RFPN module. The RPN is trained with the objectness score loss ${L}_{os}$~\cite{mohan2020efficientps} and object proposal loss ${L}_{op}$~\cite{mohan2020efficientps}

In the subsequent stage, ROI align extracts features by directly pooling from the n$^{\text{th}}$ channel of the FPN encodings with a $14\times14$ spatial resolution bounded within the object proposals obtained in the previous stage. These extracted features are then fed to specialized bounding box regression, object classification and mask segmentation networks. The second stage is trained with the classification loss ${L}_{cls}$~\cite{mohan2020efficientps}, bounding box loss ${L}_{bbx}$~\cite{mohan2020efficientps} and mask segmentation loss ${L}_{mask}$~\cite{mohan2020efficientps}.

The overall loss of the instance segmentation head is the equally weighted summation of the aforementioned losses as
\begin{equation}\label{eq:Linstance}
{L}_{instance} = {L}_{os} + {L}_{op} + {L}_{cls} + {L}_{bbx} + {L}_{mask}.
\end{equation}
Note that the gradient from the losses ${L}_{cls}$, ${L}_{bbx}$ and ${L}_{mask}$ are allowed to flow only through the network backbone and not through the RPN.

\subsection{Panoptic Fusion}

We fuse the outputs of the semantic and instance heads using the heuristic proposed in~\cite{mohan2020efficientps} to yield the panoptic predictions in the projection domain. This heuristic enables us to adaptively fuse the predictions of both the heads which alleviates the inherent overlap problem. The instance head outputs a set of object instances comprising of class prediction, confidence score, bounding box, and mask logits for each instance. While, the semantic head outputs semantic logits of $N_{stuff}+N_{thing}$ channels. We first compute the mask logit $ML_A$ for the object instances by applying a series of operations on the outputs of the instance head, consisting of thresholding, sorting, scaling, resizing, padding, and overlap filtering. Subsequently, we compute the mask logit $ML_B$ for the corresponding object instances from the outputs of the semantic head by channel selection based on the class of the object instances and suppress the logits for that channel outside the instance bounding box. Finally, we adaptively fuse the two logits $ML_A$ and $ML_B$ to yield the fused mask logits $FL$ of the instances as
\begin{equation}
\label{eq}
FL = (\sigma(ML_A) + \sigma(ML_B)) \odot (ML_A + ML_B),
\end{equation}
where $\odot$ is the Hadamard product, and $\sigma(\cdot)$ is the sigmoid function. In the next step, we concatenate the 'stuff' logits from the output of the semantic head and the fused logits, followed by applying the softmax function. Subsequently, we apply the argmax function along the channel dimension to obtain the intermediate panoptic prediction. To compute the final output, we replace the non-'thing' class predictions in the intermediate prediction with the 'stuff' class predictions of the semantic head while ignoring the classes that have an area smaller than a pre-defined area threshold $min_{sa}$.

\subsection{Panoptic Periphery Loss}

We propose the panoptic periphery loss function which exploits range information to refine the boundaries of the 'thing' class objects. By minimizing this loss, the boundary pixels of instances are adapted to maximize the range difference to the adjacent background pixels. This is motivated by the fact that there is typically a range gap at the borders of object instances. Consider the network provides a set of instances \textit{I} of 'thing' class objects, and for each instance, we have its foreground and background pixels. Then for a given range image \textit{R}, the panoptic periphery loss function is defined as
\begin{equation} 
L_{refine} =  - \frac{1}{|B|}  \sum_{i \in I}  \sum_{b \in B_i}  [\max_{n \in N}(k_n *(r_b - r_n)^2)],
\end{equation}
where $|B|$ is the total number of boundary points over all instances, $B_i$ is the set of boundary pixels for instance \textit{i}, $N$ is the set of the four immediate neighbors of pixel location $b$, $r_b$ and $r_n$ are the range value at pixel location $b$ and $n$, respectively. $k_n = 1$ for $n$ being a background pixel and $0$ otherwise. The negative sign ensures that the loss decreases when the range difference between boundary and background increases.

The overall loss $L$ for training EfficientLPS is given by
\begin{equation}\label{eq:Loverall}
{L} = {L}_{sem}+ {L}_{instance} + {L}_{refine},
\end{equation}
where $L_{sem}$ is the semantic head loss, ${L}_{instance}$ is the instance head loss and $L_{refine}$ is the panoptic periphery loss for boundary refinement.

\subsection{Back-projection}
\label{post_proc}

During the projection to point clouds, different points may get assigned to the same pixel in the projected image, which leads to the assignment of the same label to all overlapping points. Moreover, due to the downsampling operations in the network, the convolutions produce blurry outputs in the decoder which leads to leaking of the labels at the boundaries of the instances during back-projection to the 3D domain.

We use a k-nearest neighbor (kNN) based \myworries{back-projection} scheme~\cite{milioto2019rangenet++} to mitigate these issues. For every point in the point cloud, the nearest k neighbors to the point vote for its semantic and instance labels. We obtain the labels of the selected neighbors from the corresponding pixels in the projected output prediction. To compute the nearest neighbors, we search for nearest neighbors within a pre-defined window around the pixel in the projected range image, out of which we select k nearest points based on the differences in their absolute range value. The entire post-processing is GPU optimized and is only employed during inference.
    
\subsection{Psuedo Labeling}

Due to the arduous task of annotating point-wise panoptic segmentation labels in point clouds and the effectiveness of pseudo labeling in the image domain, we explore its utility for LiDAR panoptic segmentation. We formulate a novel heuristic to improve the performance of EfficientLPS without requiring any additional manual human-annotations or model augmentations. We make the following assumptions while formulating the proposed heuristic. First, the unlabeled dataset is drawn from the same data distribution as the labeled dataset. Second, the model which is used to generate the labels for the unlabeled dataset and the model learning from these generated pseudo labels are the same, i.e., both the models have the same representation capacity. Finally, the precision and recall of the model generating the pseudo labels are tunable during inference time via adjustment of one or more hyperparameters. 

\begin{figure*}
\centering
\footnotesize
{\renewcommand{\arraystretch}{1}
\begin{tabular}{P{7.5cm}P{7.5cm}}
SemanticKITTI & nuScenes \\
\\
\includegraphics[width=\linewidth]{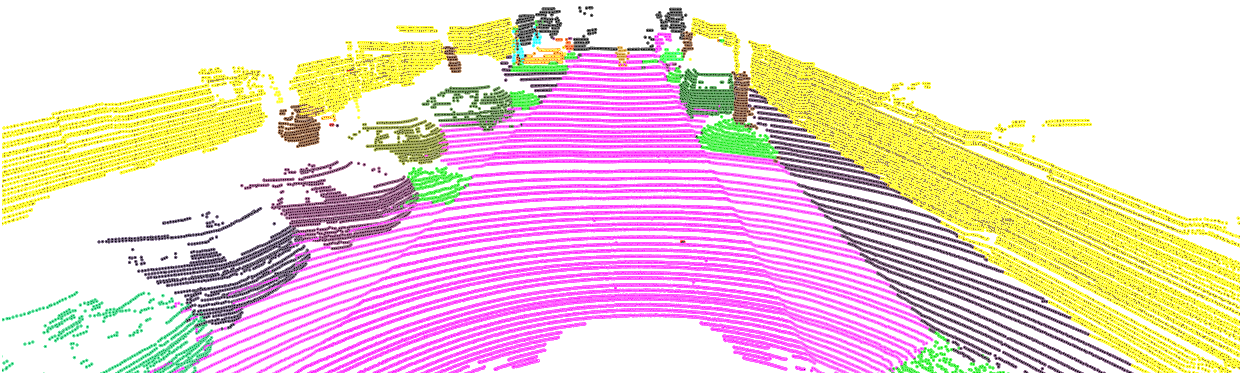} & \includegraphics[width=\linewidth]{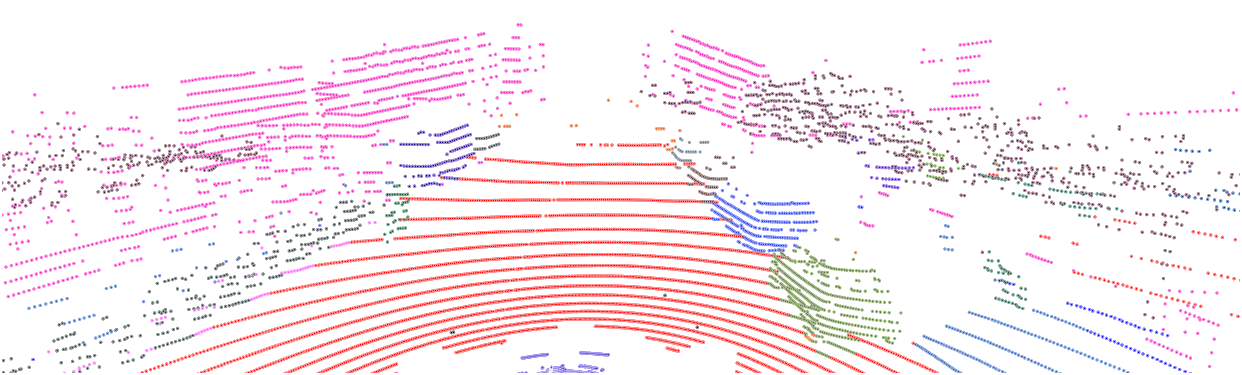} \\
\\
\includegraphics[width=\linewidth]{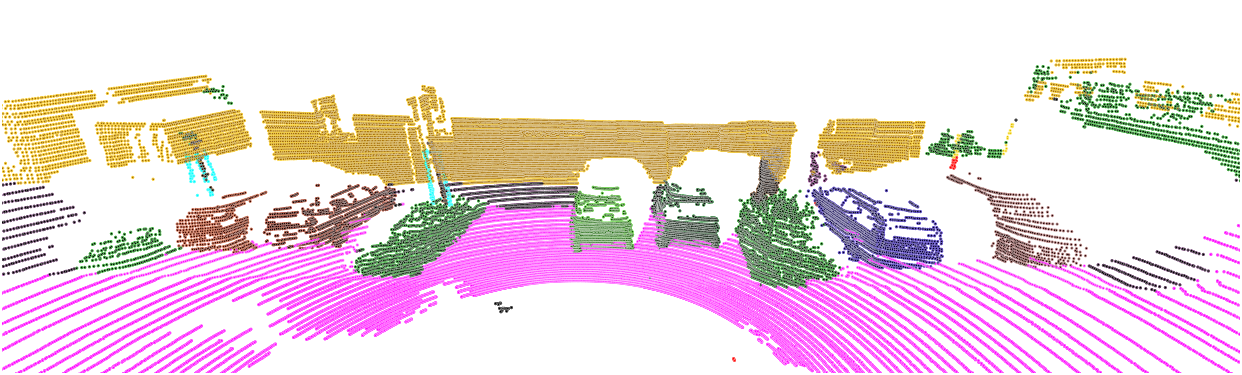} & \includegraphics[width=\linewidth]{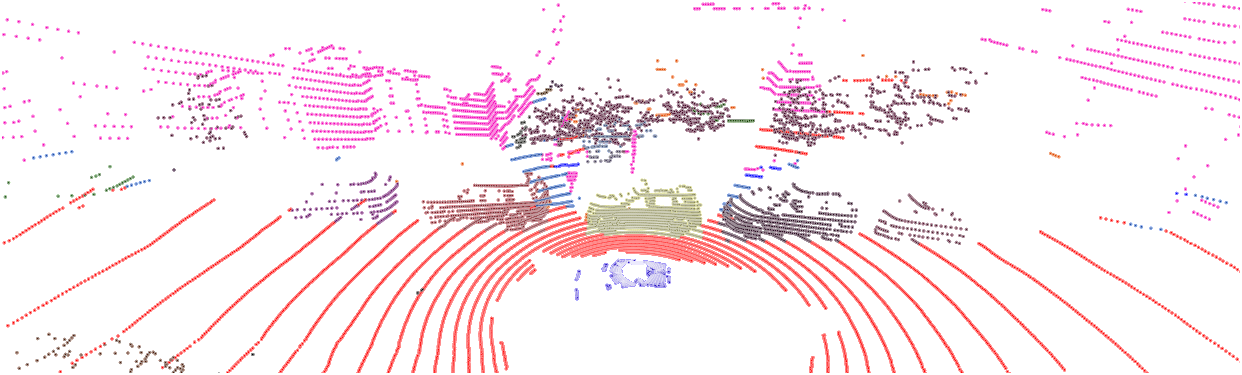} \\
\end{tabular}}
\caption[datasets visualization]{Example groundtruth visualization from SemanticKITTI and nuScenes datasets. The SemanticKITTI dataset was collected using a 64-beam LiDAR, hence it provides a fairly dense representation of the environment in comparison to the nuScenes dataset which was collected using a 32-beam LiDAR.}
\label{fig:datasets}
\end{figure*}

The first assumption is dataset-specific to ensure high-quality pseudo labels, since the model trained on the same distribution as the unseen dataset tends to generalize better than the dataset from a different distribution. In our case, we choose the KITTI RAW dataset~\cite{Geiger2013IJRR} as the unlabeled dataset since SemanticKITTI~\cite{behley2019iccv} which is the labeled dataset, is a subset of the former. We use the same EfficientLPS model to generate the pseudo labels with the goal of improving its performance, hence satisfying the second assumption. This assumption ensures that the label generating model can provide meaningful pseudo labels and the learning model has the representational capacity to capture it. To satisfy the third condition, the performance of EfficientLPS can be tuned using softmax confidence thresholding or by tuning the thresholds of the panoptic fusion module in EfficientLPS or a combination of both. This condition is required to design an effective regularization strategy for pseudo label generation in order to minimize the confirmation bias while learning.

We first train EfficientLPS on the labeled dataset and use this model to generate pseudo labels for the unlabelled dataset. We refer to this model as the Pseudo Label Generator (PLG) and the parameters that can be used to control the performance of the model as control parameters as a whole. In the next step, we use grid search to find the most optimal control parameter combination that maximizes the given ratio $(TP-FP)/TP$ until the PQ score is higher than the $PQ_{cutoff}$ parameter on the validation dataset of the labeled dataset. Here, $TP$ is the true positives, and $FP$ is false positives that are computed over the validation set. $PQ_{cutoff}$ is the minimum value of $PQ$ to which the performance of PLG is allowed to drop. By maximizing the aforementioned ratio, we make the generated pseudo label to be more accurate by having relatively fewer false positives and higher true positives. Subsequently, we use the optimal control parameter setting to generate pseudo labels with PLG. As a post-processing step, we then discard all the instances with the number of points less than a pre-defined limit $P_{limit}$ in the pseudo labels. This improves the quality of the generated pseudo labels by discarding incorrect predictions that were made due to the lack of sufficient points. We then train EfficientLPS from scratch on this pseudo labeled dataset, followed by fine-tuning the model on the labeled dataset to improve the overall performance. We comprehensively evaluate the performance of our proposed heuristic in the ablation study presented in \secref{sec:psuedolabel}.

\section{Experimental Evaluation}
\label{sec:experiments}

In this section, we first briefly describe the datasets that we report the results on in \secref{sec:dataset}, followed by the training protocol that we employ in \secref{sec:training} and detailed comparisons as well as benchmarking results in \secref{sec:comparisonSOTA}. Subsequently, we present comprehensive ablation studies on the various proposed architectural components in EfficientLPS in \secref{sec:ablation} and detailed qualitative analysis in \secref{sec:qualitative}. In all the experiments presented in this section, we use the PQ metric~\cite{kirillov2019panoptic} as the main evaluation criteria as defined by the benchmarks. For completeness, we also report the mean Intersection-over-Union (mIoU), Segmentation Quality (SQ), and Recognition Quality (RQ), as well as the aforementioned metrics for the 'stuff' and 'thing' classes separately.

\subsection{Dataset}
\label{sec:dataset}

We evaluate the performance of our approach on two datasets that were collected with LiDARs of different resolutions to test the generalization of our network. The first dataset is SemanticKITTI~\cite{behley2019iccv} which contains sequences (00-21) consisting of point-wise semantic and temporally consistent instance labels for the 43,552 LiDAR scans. The dataset is split into 20,351 scans (sequences 00-10) that are available for training, while the rest of the sequences (11-21) are withheld and used by the benchmarking server for evaluations. The dataset provides labels of 28 classes out of which 19 classes are considered for evaluation.

For the second dataset, we use nuScenes~\cite{caesar2020nuScenes} which is a large-scale dataset for autonomous driving that consists of point-wise semantic labels for 32 classes out of which 16 are considered for evaluation. The dataset contains 1000 scenes, out of which 700 scenes are used for the training set, 150 scenes for the validation set, and the rest 150 scenes for the test set. The dataset itself does not provide any panoptic segmentation labels but it contains 3D bounding box annotations for the 'thing' object classes. To obtain the panoptic segmentation labels, we extract the points that lie inside the 'thing' class bounding boxes and assign unique instance-ids. The SemanticKITTI dataset ignores object instances that have less than 50 points. We follow a similar scheme for the nuScenes dataset and ignore object instances that have less than 15 points, since nuScenes is roughly 3$\times$ sparser than SemanticKITTI. While the vertical field of view of the LiDAR that was used in nuScenes is slightly higher than SemanticKITTI, the number of vertical lines is only 32 compared to 64 in SemanticKITTI. \figref{fig:datasets} shows example groundtruth panoptic segmentation of point clouds from both these datasets. 




\subsection{Training Protocol}
\label{sec:training}

\begin{table*}
\begin{center}
\caption{Comparison of LiDAR panoptic segmentation performance on SemanticKITTI test set. All scores are in [\%].}
\label{tab:testKITTI}
\footnotesize
\begin{tabular}
{l|cccc|ccc|ccc|c}
\toprule
Method & PQ & PQ$^\dagger$  & SQ & RQ & PQ\textsuperscript{Th} & SQ\textsuperscript{Th} & RQ\textsuperscript{Th} & PQ\textsuperscript{St} & SQ\textsuperscript{St} & RQ\textsuperscript{St} & mIoU  \\
\midrule
RangeNet++~\cite{milioto2019rangenet++} + PointPillars~\cite{lang2019pointpillars} & $37.1$ &$45.9$ &  $75.9$ & $47.0$ & $20.2$ & $75.2$ & $25.2$ & $49.3$ & $76.5$ & $62.8$ &  $52.4$ \\
KPConv~\cite{thomas2019kpconv} + PointPillars~\cite{lang2019pointpillars} & $44.5$ & $52.5$ &$80.0$ & $54.4$ & $32.7$ & $81.5$ & $38.7$ & $53.1$ & $79.0$ & $65.9$ &  $58.8$ \\
LPSAD~\cite{milioto2020lidar} & $38.0$ &$47.0$ & $76.5$ & $48.2$ & $25.6$ & $76.8$ & $31.8$ & $47.1$ & $76.2$ & $60.1$ & $50.9$ \\
PanopticTrackNet~\cite{hurtado2020mopt} & $43.1$ & $50.7$ & $78.8$ & $53.9$& $28.6$ & $80.4$& $35.5$ &$53.6$ &$77.7$ & $67.3$ & $52.6$ \\
Panoster~\cite{gasperini2020panoster} & $52.7$ & $59.9$&  $80.7$ &$64.1$ & $49.4$ & $83.3$ & $58.5$ & $55.1$ & $78.8$ & $68.2$ & $59.9$ \\
\midrule
EfficientLPS (ours) & $\mathbf{57.4}$ & $\mathbf{63.2}$ & $\mathbf{83.0}$ & $\mathbf{68.7}$ & $\mathbf{53.1}$ & $\mathbf{87.8}$ & $\mathbf{60.5}$ & $\mathbf{60.5}$ & $\mathbf{79.5}$ & $\mathbf{74.6}$ & $\mathbf{61.4}$\\
\bottomrule
\end{tabular}
\end{center}
\end{table*}

\begin{table*}
\setlength\tabcolsep{3.7pt}
\begin{center}
\caption{Class-wise PQ scores on SemanticKITTI test set. R.Net, P.P, KPC refer to RangeNet++, Point Pillars, KPConv respectively. All scores are in [\%].}
         \label{tab:test_pq}
         \footnotesize
        \begin{tabular}{l|ccccccccccccccccccc|c}
            \toprule
            Method & \begin{sideways}car\end{sideways} & \begin{sideways}truck\end{sideways} & \begin{sideways}bicycle\end{sideways} & \begin{sideways}motorcycle\end{sideways} & \begin{sideways}other vehicle\end{sideways} & \begin{sideways}person\end{sideways} & \begin{sideways}bicyclist\end{sideways} & \begin{sideways}motorcyclist\end{sideways} & \begin{sideways}road\end{sideways} & \begin{sideways}sidewalk\end{sideways} & \begin{sideways}parking\end{sideways} & \begin{sideways}other ground\end{sideways} & \begin{sideways}building\end{sideways} & \begin{sideways}vegetation\end{sideways} & \begin{sideways}trunk\end{sideways} & \begin{sideways}terrain\end{sideways} & \begin{sideways}fence\end{sideways} & \begin{sideways}pole\end{sideways} & \begin{sideways}traffic sign\end{sideways} & PQ \\
            \midrule
            R.Net~\cite{milioto2019rangenet++}+ P.P.~\cite{lang2019pointpillars} & 66.9 & 6.7 & 3.1 & 16.2 & 8.8 & 14.6 & 31.8 & 13.5 & 90.6 & 63.2 & 41.3 & 6.7 & 79.2 & 71.2 & 34.6 & 37.4 & 38.2 & 32.8 & 47.4 & 37.1\\
            KPC~\cite{thomas2019kpconv} + P.P.~\cite{lang2019pointpillars} & 72.5 & 17.2 & 9.2 & 30.8 & 19.6 & 29.9 & 59.4 & 22.8 & 84.6 & 60.1 & 34.1 & 8.8 & 80.7 & 77.6 & 53.9 & 42.2 & 49.0 & 46.2 & 46.8 & 44.5\\
            LPSAD~\cite{milioto2020lidar} & 76.5 & 7.1 & 6.1 & 23.9 & 14.8 & 29.4 & 29.7 & 17.2 & 90.4 & 60.1 & 34.6 & 5.8 & 76.0 & 69.5 & 30.3 & 36.8 & 37.3 & 31.3 & 45.8 & 38.0 \\
            PanopticTrackNet~\cite{hurtado2020mopt} &70.8 & 14.4 & 17.8 & 20.9 & 27.4 &34.2 & 35.4& 7.9 & 91.2 & 66.1 & 50.3 & 10.5 & 81.8 & 75.9 & 42.0 & 44.3 & 42.9 & 33.4 & 51.1 & 43.1 \\
            Panoster~\cite{gasperini2020panoster} &  84.0 & 18.5 & 36.4 & 44.7 & 30.1 & 61.1 & \textbf{69.2} & $\mathbf{51.1}$ & 90.2 & 62.5 & 34.5 & 6.1 & 82.0 & 77.7 & $\mathbf{55.7}$ & 41.2 & 48.0 & $\mathbf{48.9}$ & 59.8 & 52.7\\
            \midrule
            EfficientLPS (ours) & $\mathbf{85.7}$ & $\mathbf{30.3}$ & $\mathbf{37.2}$ & $\mathbf{47.7}$ & $\mathbf{43.2}$ & $\mathbf{70.1}$ & 66.0 & 44.7 & $\mathbf{91.1}$ & $\mathbf{71.1}$ & $\mathbf{55.3}$ & $\mathbf{16.3}$ & $\mathbf{87.9}$ & $\mathbf{80.6}$ & 52.4 & $\mathbf{47.1}$ & $\mathbf{53.0}$ & 48.8 & $\mathbf{61.6}$ & $\mathbf{57.4}$ \\
            \bottomrule
        \end{tabular}
    \end{center}
\end{table*}

We train our network on projected point clouds of $4096\times256$ resolution. We use bilinear interpolation on the projections obtained from scan unfolding and nearest neighbor interpolation on the ground truth point clouds. We initialize the main encoder of our EfficientLPS architecture with weights from the EfficientNet-B5 model pre-trained on the ImageNet~\cite{deng2009imagenet} dataset. Furthermore, we initialize the weights of the iABNsync layers to 1 and use Xavier initialization for the other layers. We also employ zero constant initialization for the biases and set the slope of Leaky ReLU to 0.01. We use the same hyperparameters for the instance head as Mask~R-CNN~\cite{he2017mask}. For the panoptic fusion module, we set $c_t = 0.5$, $o_t = 0.5$ and $min_{sa} = 128$.

We use Stochastic Gradient Descent (SGD) with a momentum of $0.9$ for training our models. We employ a multi-step learning rate schedule, where we start with an initial base learning rate of $0.07$ and reduce it by a factor of $10$ after 16,000 and 22,000 iterations. We train our models for a total of 25,000 iterations with a batch size of 16 on 8 NVIDIA TITAN RTX GPUs. We first train the model on the pseudo labeled dataset until the first reduction in learning rate by a factor of $10$, followed by continuing the training on the labeled dataset.

\subsection{Comparison with the State-of-the-Art} 
\label{sec:comparisonSOTA}

In this section, we evaluate the performance of EfficientLPS and compare with state-of-the-art methods for panoptic segmentation of LiDAR point clouds. 

\noindent{\textbf{SemanticKITTI:}} We compare with three state-of-the-art methods, LPSAD~\cite{milioto2020lidar}, PanopticTrackNet~\cite{hurtado2020mopt}, and Panoster~\cite{gasperini2020panoster}, as well as the two baselines, (RangeNet++~\cite{milioto2019rangenet++} + PointPillars~\cite{lang2019pointpillars}), and (KPConv~\cite{thomas2019kpconv} + PointPillars~\cite{lang2019pointpillars}). \tabref{tab:testKITTI} presents the results on the SemanticKITTI test set which was evaluated by the benchmark server. Our proposed EfficientLPS achieves a PQ score of $57.4\%$, which is an improvement of $4.7\%$ over the previous state of the art Panoster. EfficientLPS also outperforms all the existing methods in all the metrics and sets the new state-of-the-art on this benchmark. The higher overall SQ score of EfficientLPS can be primarily attributed to the proposed panoptic periphery loss function, which improves the segmentation quality of 'thing' class objects by refining their boundaries. This is evident from the increase of $4.5\%$ in SQ$^{th}$ score compared to Panoster. Moreover, the proposed distance-dependent semantic head enables the recognition of objects in a scale-invariant manner by incorporating range encoded information to achieve a higher recognition quality, especially for the 'stuff' classes. This yields an improvement of $6.4\%$ in the RQ$^{st}$ score, which enables it to achieve the best overall RQ score of $68.7\%$. Additionally, the backbone of EfficientLPS equipped with the proposed PCM module which models the geometric transformations of different objects and the RFPN which learns spatially consistent features, contributes to the improvement of $3.7\%$ in PQ$^{Th}$, $5.4\%$ in PQ$^{St}$ and $1.5\%$ in mIoU scores.

\begin{table*}
\begin{center}
\caption{Comparison of LiDAR panoptic segmentation performance on SemanticKITTI validation set. All scores are in [\%].}
\label{tab:validationKITTI}
\footnotesize
\begin{tabular}
{l|cccc|ccc|ccc|c}
\toprule
Method & PQ & PQ$^\dagger$  & SQ & RQ & PQ\textsuperscript{Th} & SQ\textsuperscript{Th} & RQ\textsuperscript{Th} & PQ\textsuperscript{St} & SQ\textsuperscript{St} & RQ\textsuperscript{St} & mIoU  \\
\midrule
RangeNet++~\cite{milioto2019rangenet++} + PointPillars~\cite{lang2019pointpillars} & $36.5$ & - &  $73.0$ & $44.9$ & $19.6$ & $69.2$ & $24.9$ & $47.1$ & $75.8$ & $59.4$ &  $52.8$ \\
KPConv~\cite{thomas2019kpconv} + PointPillars~\cite{lang2019pointpillars} & $41.1$ & - &$74.3$ & $50.3$ & $28.9$ & $69.8$ & $33.1$ & $50.1$ & $77.6$ & $62.8$ &  $56.6$ \\
LPSAD \cite{milioto2020lidar} & $36.5$ & $46.1$ & - & - & - & - & $28.2$ & - & - & - & $50.7$ \\
PanopticTrackNet~\cite{hurtado2020mopt} & $40.0$ & - & $73.0$ & $48.3$ & $29.9$ & $76.8$ & $33.6$ & $47.4$ & $70.3$ & $59.1$  & $53.8$  \\
Panoster~\cite{gasperini2020panoster} & $55.6$ & - &  $79.9$ &$66.8$ & $56.6$ & - & $65.8$ & - & - & - & $61.1$ \\
\midrule
EfficientLPS (ours) & $\mathbf{59.2}$& $\mathbf{65.1}$ & $\mathbf{75.0}$ & $\mathbf{69.8}$ & $\mathbf{58.0}$ & $\mathbf{78.0}$& $\mathbf{68.2}$ & $\mathbf{60.9}$ &$\mathbf{72.8}$ & $\mathbf{71.0}$ & $\mathbf{64.9}$\\
\bottomrule
\end{tabular}
\end{center}
\end{table*}

In \tabref{tab:test_pq}, we present a comparison of the class-wise PQ scores on the SemanticKITTI test set. EfficientLPS achieves the highest PQ score for most of the 'stuff' class objects, with the exception of \textit{trunk} and \textit{pole} classes which are outperformed by Panoster. This shows that incorporating range encoded features into the semantic head helps achieve better overall semantic segmentation performance. The resulting spatial awareness is significantly beneficial for the \textit{other-ground} class, which every method struggles to classify due to the presence of the more dominating \textit{road} and \textit{sidewalk} classes. We also observe a similar effect with the \textit{parking} class. EfficientLPS achieves an improvement in the PQ score by $20.8\%$ for \textit{parking} and $10.2\%$ for \textit{other-ground} in comparison to Panoster. EfficientLPS also achieves the best performance for all 'thing' class objects, with the exception of \textit{motocyclist} and \textit{bicyclist} classes. This is due to the fact that both these classes share almost the same properties and they are relatively small objects which are represented by only a few points. These objects also always coexists with other classes such as \textit{bicycle} and \textit{motorcycle}, and are adversely affected during the projection of point cloud, rendering them very close to other objects. Hence, the point-based backbone KPConv and the clustering used by Panoster provides an advantage in this case. Nevertheless, EfficientLPS outperforms the other methods in the PQ score for all the other classes with large margins of $12.1\%$ for \textit{other vehicles}, $11.8\%$ for \textit{truck} and $9\%$ for \textit{person} classes. Therefore, the overall state-of-the-art performance obtained from our network is not just due to the improvement in scores for a particular object class, rather is a result of collective improvement across different semantic object classes with a variety of structural properties.

\tabref{tab:validationKITTI} presents the results on the SemanticKITTI validation set. `-' indicates that the corresponding methods do not report the specific metric. We observe a similar trend here as the test set where EfficientLPS outperforms all the other methods in all of the metrics. It achieves a PQ score of $59.2\%$ and $75.0\%$, $69.8\%$ and $64.9\%$ for SQ, RQ and mIoU scores, respectively. 

\noindent{\textbf{nuScenes:}} 
As there are no existing panoptic segmentation methods that have been benchmarked on the nuScenes dataset, we trained two baseline methods by combining individual semantic and instance segmentation models, namely (KPConv~\cite{thomas2019kpconv} + Mask~R-CNN~\cite{he2017mask}) and (RangeNet++~\cite{milioto2019rangenet++} + Mask~R-CNN~\cite{he2017mask}), as well as the established PanopticTrackNet\cite{hurtado2020mopt} which is a projection-based panoptic segmentation model for LiDAR point clouds. For all these models, we use the original code provided by the authors and optimized the hyperparameters to the best of our ability. We trained Mask~R-CNN on the projected point cloud images using the approach described in \secref{sec:scanUnfold} and project the predictions back to the 3D domain using the post-processing described in \secref{post_proc}. We also make these trained baselines publicly available.

\begin{table*}
\begin{center}
\caption{Comparison of LiDAR panoptic segmentation performance results on nuScenes validation set. All scores are in [\%].}
\label{tab:Nuscenes}
\footnotesize
\begin{tabular}
{l|cccc|ccc|ccc|c}
\toprule
Method & PQ & PQ$^\dagger$  & SQ & RQ & PQ\textsuperscript{Th} & SQ\textsuperscript{Th} & RQ\textsuperscript{Th} & PQ\textsuperscript{St} & SQ\textsuperscript{St} & RQ\textsuperscript{St} & mIoU  \\
\midrule
RangeNet++ \cite{milioto2019rangenet++} + Mask R-CNN \cite{he2017mask}& 46.6 & 52.6 & 79.5 & 58.4 & 39.9 & 80.5& 52.1 & 57.8 & 77.9 & 68.8 & 56.6 \\
PanopticTrackNet~\cite{hurtado2020mopt} & 51.4 & 56.2 &80.2 & 63.3 & 45.8 &81.4 &55.9 & 60.4 & 78.3 & 75.5 & 58.0 \\
KPConv \cite{thomas2019kpconv} + Mask R-CNN \cite{he2017mask}& 51.5 & 56.8 &80.3 & 63.5 & 44.6 &81.3 &53.9 & 62.9 & 78.8 & 79.4 & 58.9 \\
\midrule
EfficientLPS (ours) & \textbf{62.0} & \textbf{65.6} & \textbf{83.4} & \textbf{73.9} & \textbf{56.8} & \textbf{83.2} & \textbf{68.0}& \textbf{70.6}  & \textbf{83.8} & \textbf{83.6} & \textbf{65.6} \\
\bottomrule
\end{tabular}
\end{center}
\end{table*}

\begin{table*}
\setlength\tabcolsep{3.7pt}
\begin{center}
\caption{Class-wise results on nuScenes validation set. All scores are in [\%].} 
\label{tab:nuScenesClass}
\begin{tabular}{l|cccccccccccccccc|c}
\toprule
Method & \begin{sideways}barrier\end{sideways} & \begin{sideways}bicycle\end{sideways} & \begin{sideways}bus\end{sideways} & \begin{sideways}car\end{sideways} & \begin{sideways}cvehicle\end{sideways} & \begin{sideways}motorcycle\end{sideways} & \begin{sideways}pedestrian\end{sideways} & \begin{sideways}traffic cone\end{sideways} & \begin{sideways}trailer\end{sideways} & \begin{sideways}truck\end{sideways} & \begin{sideways}driveable\end{sideways} & \begin{sideways}other flat\end{sideways} & \begin{sideways}sidewalk\end{sideways} & \begin{sideways}terrain\end{sideways} & \begin{sideways}man-made\end{sideways} & \begin{sideways}vegetation\end{sideways}  & PQ \\
\midrule
RangeNet++~\cite{milioto2019rangenet++} + Mask R-CNN~\cite{he2017mask} & 40.3 & 25.7 & 51.7 & 62.5 & 14.6 & 48.3 & 38.8 & 41.8 & 32.7 & 43.0 & 77.1 & 41.5 & 59.2 & 42.1 & 58.9 & 67.9 & 46.6 \\
PanopticTrackNet~\cite{hurtado2020mopt} & 47.1 & 32.9 & 57.9 & 66.3 & 22.8 & 51.1 & 42.8 & 46.8 & 38.9 & 51.0 & 81.5 & 42.3 & 61.8 & 45.1 & 60.9 & 70.9 & 51.4  \\
KPConv~\cite{thomas2019kpconv} + Mask R-CNN~\cite{he2017mask} & 46.7 & 31.5 & 56.8 & 65.7 & 21.9 & 50.4 & 41.6 & 44.9 & 37.6 & 49.1 & 83.5 & 43.1 & 63.5 & 48.6 & 73.9 & 71.5 & 51.5  \\
\midrule
EfficientLPS (ours) & \textbf{56.8} & \textbf{37.8} & \textbf{52.4} & \textbf{75.6} & \textbf{32.1} & \textbf{65.1} & \textbf{74.9} & \textbf{73.5} & \textbf{49.9} & \textbf{49.7} & \textbf{95.2} & \textbf{43.9} & \textbf{67.5} & \textbf{52.8} & \textbf{81.8} & \textbf{82.4} & \textbf{62.0}  \\
\bottomrule
\end{tabular}
\end{center}
\end{table*}

\tabref{tab:Nuscenes} presents the results on the nuScenes validation set. Among the baselines, (KPConv + Mask~R-CNN) achieves the highest PQ score of $51.5\%$, closely followed by PanopticTrackNet which achieves a PQ score of $51.3\%$. (KPConv + Mask~R-CNN) achieves a higher PQ$^{St}$ score than PanopticTrackNet which demonstrates that ability of point-based methods to perform better at semantic segmentation. During the projection of points, distant objects in the 3D domain end up close to each other in the projected 2D domain. Hence, projection based methods that solely operate in the 2D domain find it hard to distinguish between them. The proposed distance-dependent semantic head in EfficientLPS exploits range encoded features and achieves an improvement of $7.7\%$ in the PQ$^{St}$ score over (KPConv + Mask~R-CNN). On the other hand, the top-down architecture of PanopticTrackNet which has a dedicated instance segmentation head, achieves a better performance in segmenting instances of 'thing' class objects, thereby achieving a higher PQ$^{Th}$ score than (KPConv + Mask~R-CNN). The RFPN moodule along with proposed panoptic periphery loss that we use for training EfficientLPS enables it to achieve an improvement of $12.2\%$ in the PQ$^{Th}$ score over PanopticTrackNet. Overall, EfficientLPS achieves  a PQ score of $62.0\%$, SQ score of $83.4\%$, RQ score of $73.9\%$, and mIoU of $65.6\%$, outperforming all the methods in each of the metrics and sets the new state of the art on the nuScenes dataset. The consistent state-of-the-art performance, even on the sparse nuScenes dataset demonstrates the effectiveness and the generalization ability of our proposed modules in tackling different challenges such as distance-dependent sparsity, severe occlusions, large scale-variations, and re-projection errors.

\begin{table*}
\centering
\caption{Ablative analysis on the various proposed architectural components in EfficientLPS. The model variants consists of the marked (\cmark) modules in their respective columns with Per. Loss denoting the Panoptic  and P. Labels denoting  Labels. The results are reported on the SemanticKITTI validation set.} 
\label{tab:ablation}
\begin{tabular}{@{}c|ccccc|ccc|c|c|c|c@{}}
\toprule
Model & PCM & RFPN & RDPC & Panoptic & Pseudo & PQ  & SQ   & RQ  & PQ\textsuperscript{St}&
PQ\textsuperscript{Th} & mIoU & \myworries{Runtime} \\
Variant & & & & Periphery Loss & Labels & (\%) & (\%) & (\%) & (\%) & (\%) & (\%) & \myworries{$(\si{\milli\second})$}\\
\midrule
M1 & \xmark & \xmark  & \xmark & \xmark & \xmark & $53.0$& $73.1$  & $63.9$ &  $53.3$ &$52.5$  & $58.6$ &\myworries{$153.84$} \\ 
M2 & \cmark & \xmark  & \xmark & \xmark  & \xmark &$53.9$& $73.9$  & $64.4$ & $54.3$&$53.3$ &$59.8$ &
\myworries{$175.43$}\\
M3 & \cmark & \cmark  & \xmark & \xmark & \xmark&  $55.7$& $74.4$  & $65.8$ &$55.7$ &$55.8$&  $60.9$ & \myworries{$192.31$}  \\
M4 & \cmark & \cmark  & \cmark & \xmark& \xmark &  $56.6$& 75.0  & $66.8$ &$56.4$ &$56.9$ & $62.4$ & \myworries{$212.76$}   \\
M5 & \cmark & \cmark  & \cmark & \cmark & \xmark& $57.4$& 75.0  & $67.6$ &$56.5$ &$58.7$ & $62.5$ & \myworries{$212.76$}   \\
M6 & \cmark & \cmark  & \cmark & \cmark & \cmark& \textbf{59.2} & \textbf{75.0} & \textbf{69.8} & \textbf{58.0} & \textbf{60.9} & \textbf{64.9} & \myworries{$212.76$} \\
\bottomrule
\end{tabular}
\end{table*}

In \tabref{tab:nuScenesClass}, we present the class-wise PQ scores on the nuScenes validation set. The sparsity of the point clouds especially affects segmentation of smaller objects such as \textit{pedestrian}, \textit{bicycle}, \textit{motorcycle}, \textit{barrier} and \textit{traffic cone}. The operation of the proposed proximity convolution module is independent of the sparsity of the point cloud, and models the geometric transformations of objects, only based on the proximity between the points. This is one of the key factors that enables EfficientLPS to achieve a higher performance for all 'thing' object classes, with an improvement of $12.9\%$ for \textit{pedestrian}, $8.4\%$ for \textit{bicycle}, and $13.2\%$ for \textit{motorcycle} compared to PanopticTrackNet. The sparse nature of point clouds also makes it harder to recognize and distinguish between different semantic object classes, especially for the object classes that often appear close to each other such as \textit{driveable surface}, \textit{sidewalk} and \textit{terrain}. The explicit incorporation of range encoded information in RFPN and the distance-dependent semantic head introduces spatial consistency in the learned features. This especially helps 'stuff' object classes, leading to an PQ improvement by $14.0\%$ for \textit{driveable surface}, $7.2\%$ for \textit{terrain}, and $7.0\%$ for \textit{sidewalk} compared to (KPConv + Mask~R-CNN). Overall, the superior performance obtained for all classes demonstrates the efficacy of EfficientLPS for accurately segmenting different semantic object classes, even for very sparse point clouds.

\subsection{Ablation Studies}
\label{sec:ablation}

In this section, we present extensive ablation studies on the various proposed architectural components in EfficientLPS to warrant our design choices and study the impact of each module on the overall performance of our architecture. We begin with a comprehensive analysis on the EfficientLPS architecture that shows the effect of our proposed proximity input layer, range-aware FPN, distance-dependent semantic head, panoptic periphery loss, and pseudo labeling framework, on the overall performance of our network. We then study the influence of parameters for the search grid and kernel size on the performance of the proximity convolution module. Subsequently, we study the effect of the REN and the different approaches to incorporate it with the 2-way FPN in our proposed range-aware FPN. We then present detailed analysis of the distance-dependent semantic head to show the impact of different architectural design choices. Furthermore, we quantitatively show the boundary refinement achieved with the panoptic periphery loss function using the border IoU metric and compare the performance of the proposed pseudo labeling heuristic to a naive heuristic. Finally, to demonstrate the generalization ability of our proposed modules, we present results by incorporating our architectural components into other well-known top-down panoptic segmentation networks in a straight forward manner.

\subsubsection{Comprehensive Analysis of EfficientLPS}
\label{sec:detailedAnalysis}

\begin{figure}
\centering
\includegraphics[width=0.47\textwidth]{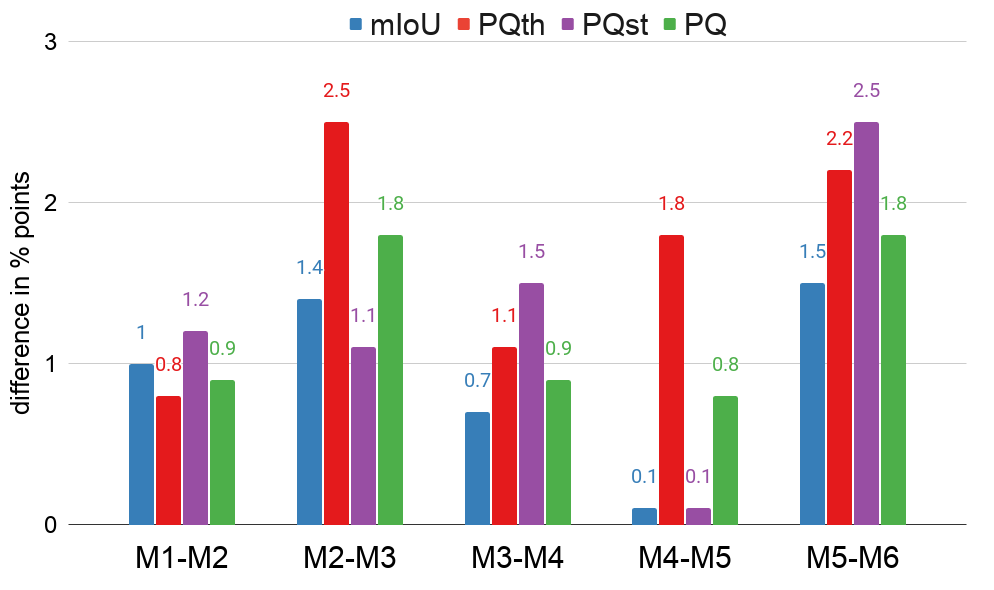}
\caption{Ablation analysis on the percent change in the metrics for the incorporation of various architectural components shown in \tabref{tab:ablation}. Where M(n)-M(n+1) denotes the \% improvement in the metrics that model M(n+1) achieves over model M(n).}
\label{fig:ablation}
\end{figure}

\begin{figure*}
\centering
\footnotesize
{\renewcommand{\arraystretch}{1}
\begin{tabular}{P{0.1cm}P{5.5cm}P{5.5cm}P{5.5cm}}
&  \raisebox{-0.4\height}{Ground Truth} &  \raisebox{-0.4\height}{Model MX}&  \raisebox{-0.4\height}{Model MY} \\
{\rotatebox[origin=c]{90}{(a) X=1, Y=2}} &\raisebox{-0.4\height}{\includegraphics[width=\linewidth]{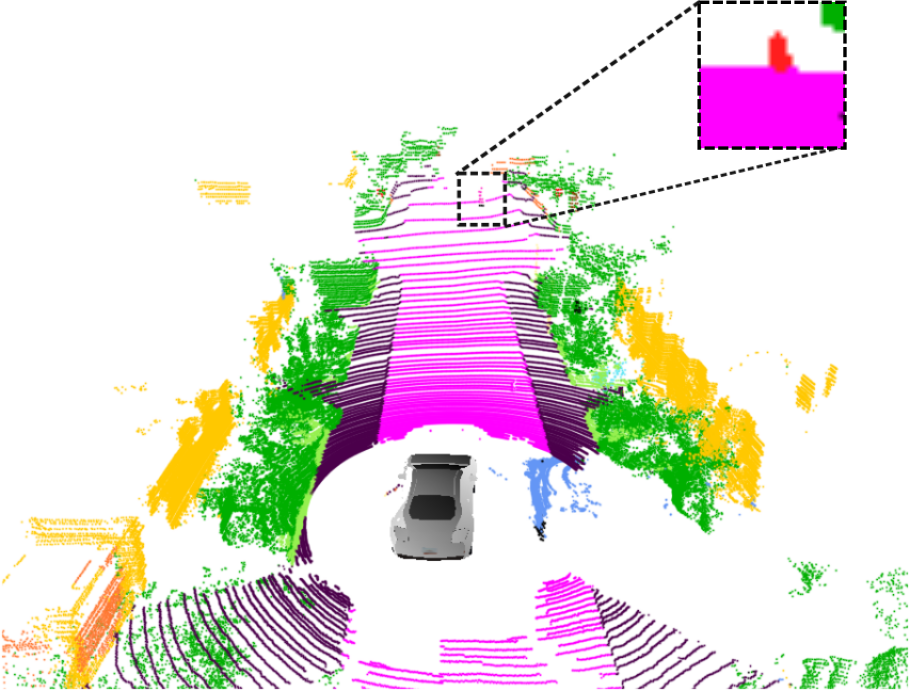}} & \raisebox{-0.4\height}{\includegraphics[width=\linewidth]{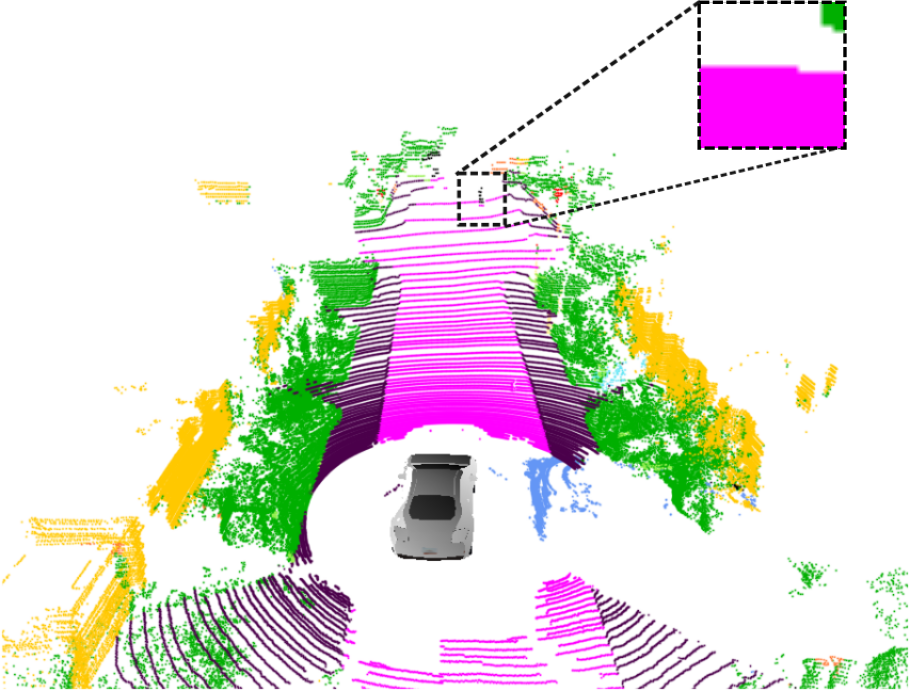}} & \raisebox{-0.4\height}{\includegraphics[width=\linewidth]{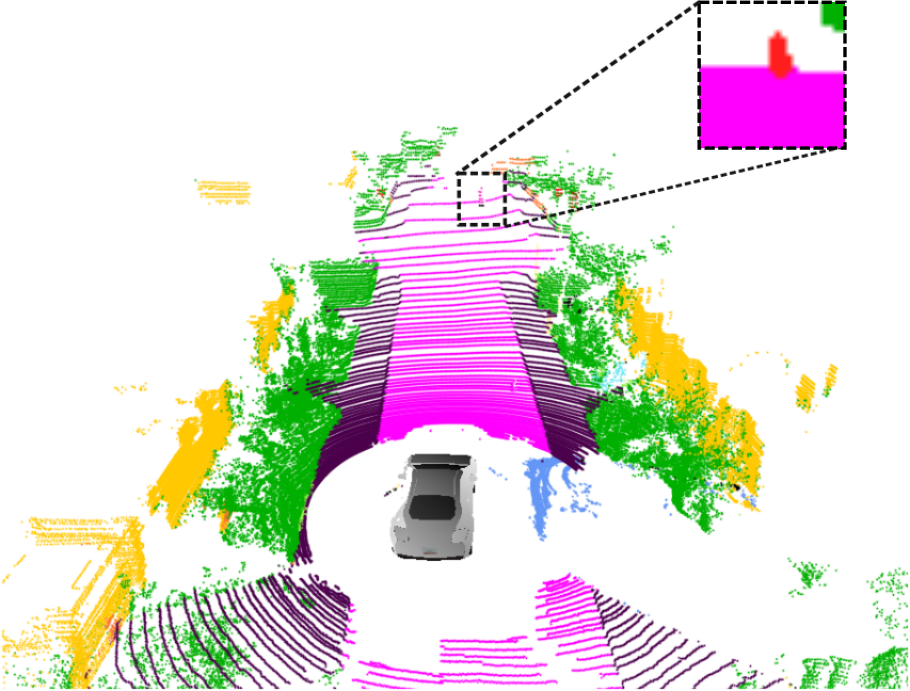}} \\
\\
{\rotatebox[origin=c]{90}{(b) X=2, Y=3}} &\raisebox{-0.4\height}{\includegraphics[width=\linewidth]{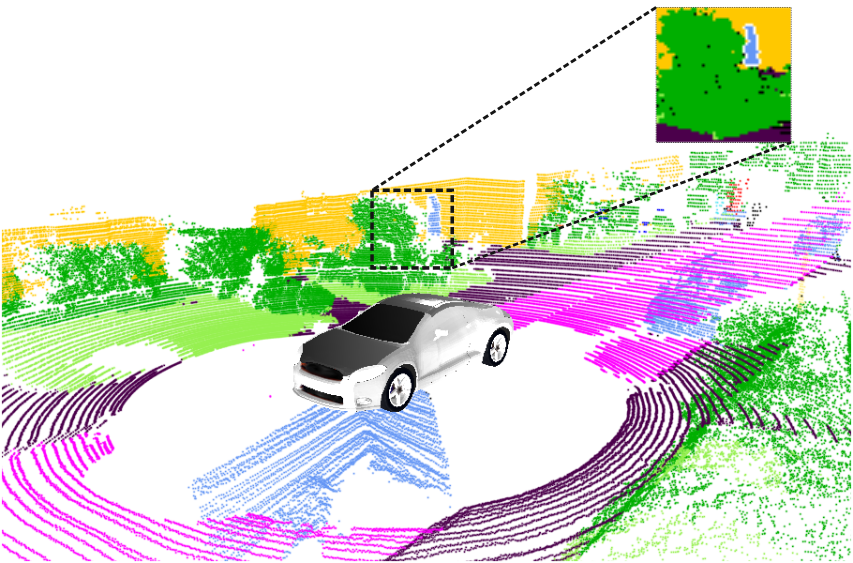}} & \raisebox{-0.4\height}{\includegraphics[width=\linewidth]{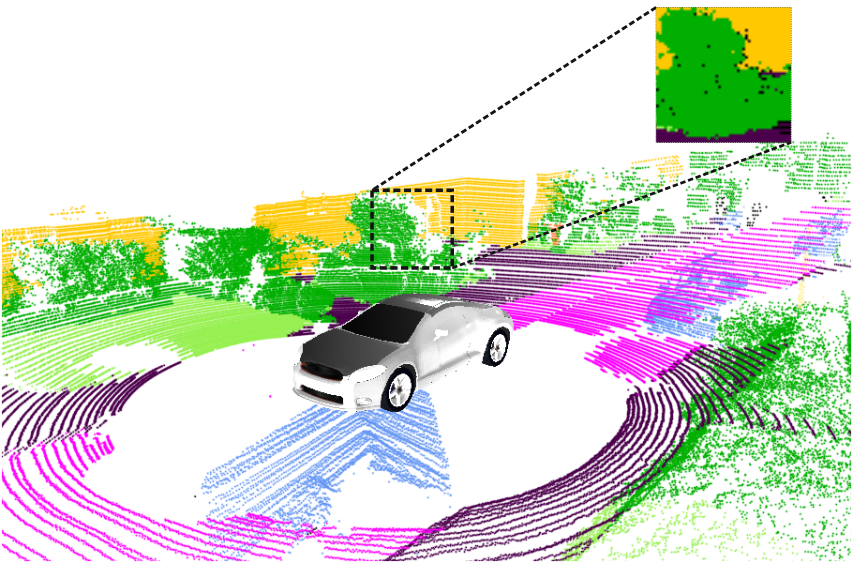}} & \raisebox{-0.4\height}{\includegraphics[width=\linewidth]{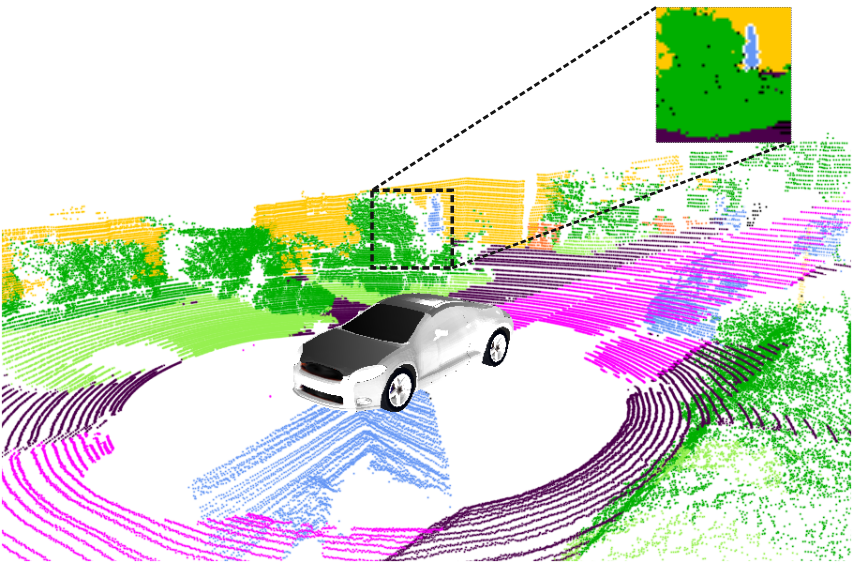}} \\
\\
{\rotatebox[origin=c]{90}{(c) X=3, Y=4}} &\raisebox{-0.4\height}{\includegraphics[width=\linewidth]{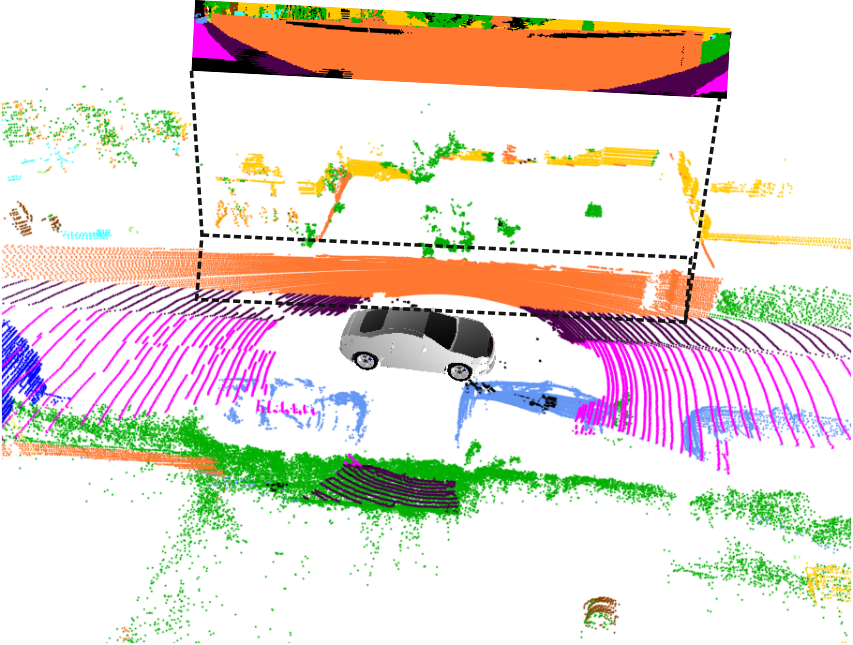}} & \raisebox{-0.4\height}{\includegraphics[width=\linewidth]{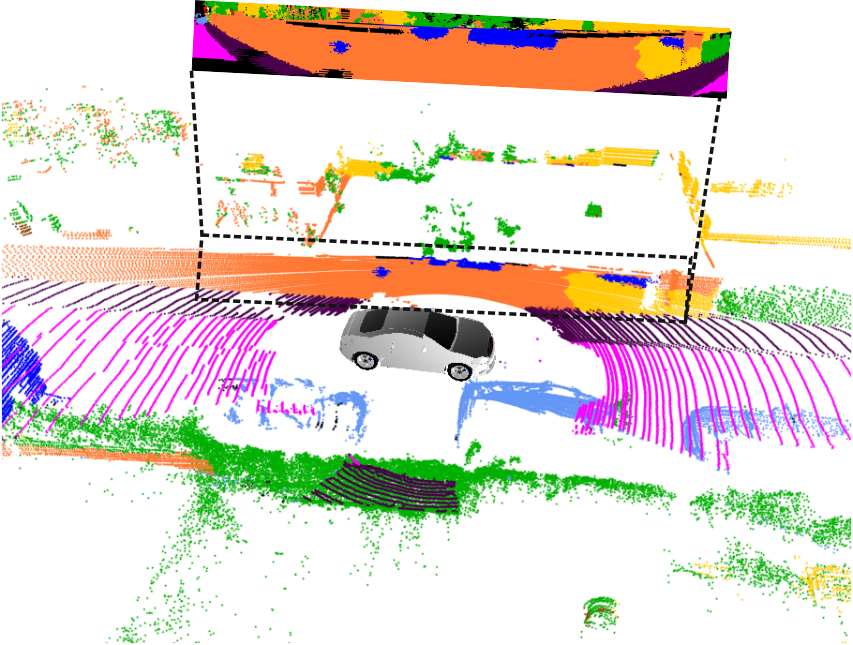}} & \raisebox{-0.4\height}{\includegraphics[width=\linewidth]{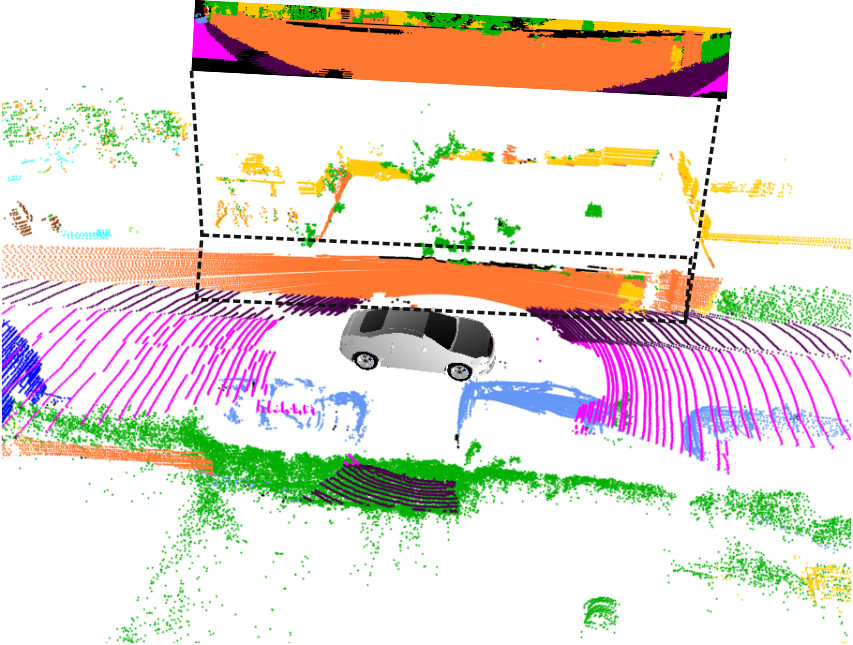}} \\
\\
{\rotatebox[origin=c]{90}{(d) X=4, Y=5}} &\raisebox{-0.4\height}{\includegraphics[width=\linewidth]{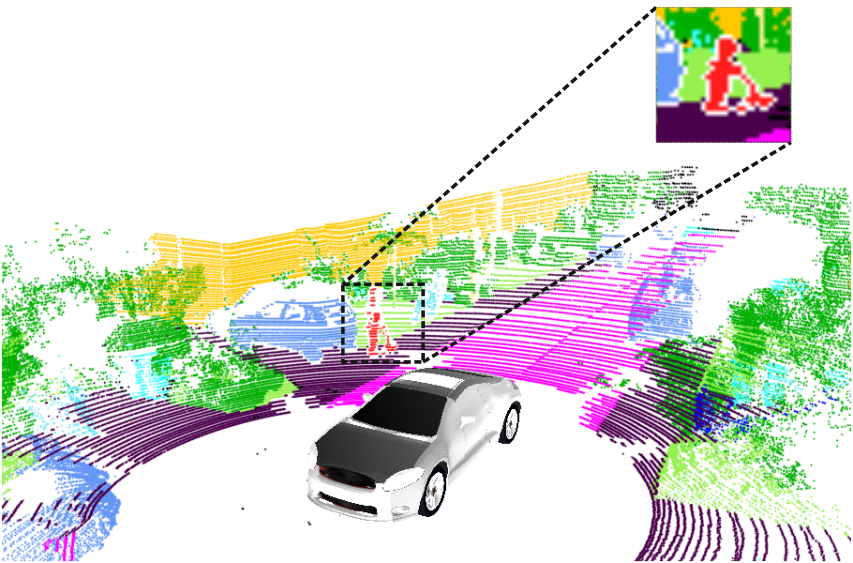}} & \raisebox{-0.4\height}{\includegraphics[width=\linewidth]{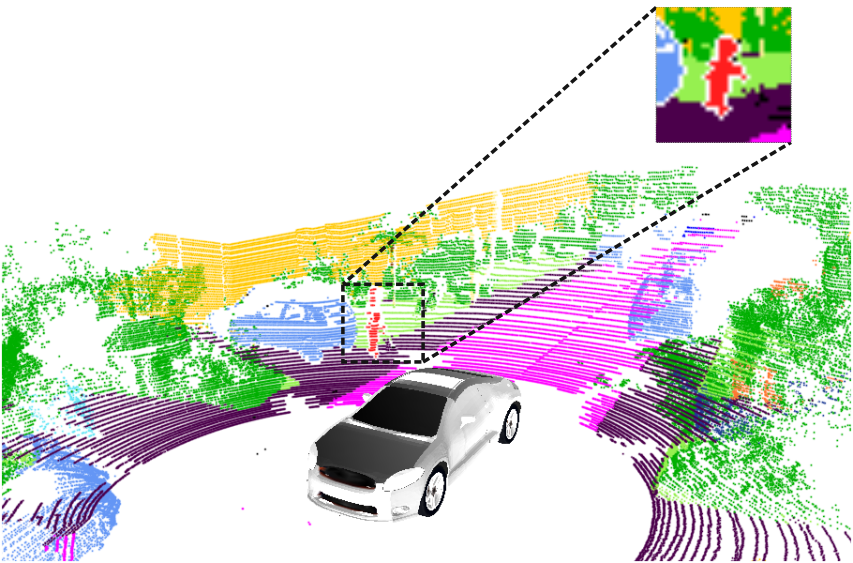}} & \raisebox{-0.4\height}{\includegraphics[width=\linewidth]{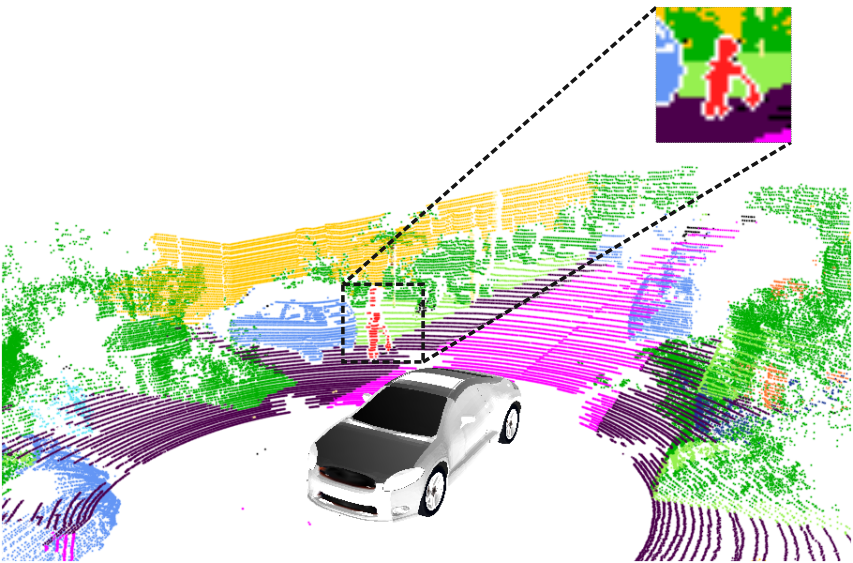}} \\
\end{tabular}}
\caption{Qualitative comparison of different models described in \tabref{tab:ablation}. The model numbers denoted by $X$ and $Y$ are shown in the first column. In Fig~(a), the incorporation of the proximity convolution module in model $M2$ enables accurate segmentation of small and distant objects such as the motorcyclist in the image. Fig.~(b) compares model $M2$ with $M3$ which incorporates the Range-aware FPN, enabling it to segment occluded objects such as the van in blue which is occluded by the surrounding vegetation. Fig.~(c) shows the performance improvement due to the new distance guided semantic head incorporated into model $M4$ which enables consistent segmentation of the sidewall shown in orange. Lastly, Fig.~(d) shows the improvement due to the panoptic periphery loss used in training model $M5$ which enables segmenting the entire instance of the human shown in red. \myworries{The enlarged images are taken from the corresponding projected range image for better visualization (PDF best viewed at $\times4$ zoom scale).}}
\label{fig:visual_ablation}
\end{figure*}

In this section, we study the improvement due to the incorporation of various architectural components proposed in EfficientLPS. The results of this experiment are shown in \tabref{tab:ablation}. \figref{fig:ablation} also shows the improvement in performance for each of the models described in this section. We begin with the base model M1 that consists of EfficientNet-B5 followed by the 2-way FPN as the shared backbone with the semantic head from \cite{mohan2020efficientps} and a Mask~R-CNN based instance head. Subsequently, the logits from both heads are fused in the panoptic fusion module at inference time. We train this model with an input resolution of $4096\times256$ as it allows the anchor scales defined in~\cite{he2017mask} to be used directly. We use scan unfolding for point cloud projection and the kNN-based post-processing for re-projecting the predicted labels back to the 3D domain. Further, we employ the \ls loss in addition to weight per-pixel log loss while training. This model M1 achieves a $PQ$ score of $53.0\%$ with $PQ^{st}$ and $PQ^{th}$ scores of $53.3\%$ and $52.5\%$ respectively. \myworries{This model has a runtime of $\SI{153.84}{\milli\second}$. To compute the runtime, we use a single NVIDIA Titan RTX GPU and an Intel Xenon@2.20GHz CPU. We average over 1000 runs on the same LiDAR point cloud. In the case of parallel components in the architecture, maximum runtime among all the components contribute to the total runtime.} In the subsequent model M2, we incorporate our proposed proximity convolution module which achieves a $PQ$ score of $53.9\%$ and an $mIoU$ of $59.8\%$, which constitutes to an improvement $0.9\%$ and $1.2\%$ respectively over the model M1. This improvement can be attributed to the enhanced transformation modeling capability imparted due to the incorporation of the proximity convolution module, as shown qualitatively in \figref{fig:visual_ablation}~(a). The model M1 fails to recognize and segment far-away objects (motorcyclist in the figure) as the points become more sparse with increasing distance. In contrast, the model M2 is able to accurately capture the distant objects. \myworries{Additionally, M2 has a runtime of $\SI{175.43}{\milli\second}$}.

The model M3 builds upon the model M2 by incorporating the range-aware FPN. This model achieves an improvement of $1.8\%$ in $PQ$ score that constitutes to an improvement of $1.4\%$ in the $PQ^{St}$ and $2.5\%$ in the $PQ^{Th}$ scores \myworries{while having a runtime of $\SI{192.31}{\milli\second}$}. This performance improvement can be attributed to the distance-aware reinforcement of coherently aggregated fine and contextual features that enables accurate segmentation of small occluded objects. In \figref{fig:visual_ablation}~(b), the car is occluded by the vegetation. The model M3 is able to successfully segment the occluded classes due to the incorporation of range-aware features, whereas model M2 falsely predicts both the occluded and the occluder objects as either the foreground or the background class. The next model M4 which incorporates our distance-dependent semantic head into model M3 achieves an improvement of $1.5\%$ in $mIoU$ and $0.9\%$ in the $PQ$ score \myworries{with runtime of $\SI{212.76}{\milli\second}$}. In order to visualize the qualitative improvement of model M4 over model M3, we present an example in \figref{fig:visual_ablation}~(c) where the model M4 segments the \textit{wall} class much more accurately than the model M3. Our proposed semantic head effectively handles the scale variation depicted in the examples. It benefits from the adaptable receptive field which results from combining fixed multi-dilation rate convolutions with range-guided depth-wise atrous separable convolutions.

Subsequently, the model M5 extends model M4 by using our proposed panoptic periphery loss during training. This model achieves $PQ^{th}$ score of $58.7\%$ which is a substantial gain of $1.8\%$ over model M4. The model $M5$ achieves a overall $PQ$ and $mIoU$ scores of $57.4\%$ and $62.5\%$ respectively. \figref{fig:visual_ablation}~(d) shows a qualitative comparison of the boundary refinement improvement. We further improve the performance of our network in model M6 by employing our pseudo labeling framework as described in \secref{sec:psuedolabel}. Training with an unlabeled dataset in combination with the labeled dataset leads to a large improvement of $1.5\%$ in $mIoU$ and $1.8\%$ in $PQ$ scores. The final model M6 achieves state of the art performance on SemanticKITTI, with a $PQ$ score of $59.2\%$. \myworries{Moreover, we do not observe any changes in runtime of models M5 and M6 since there is no additional architectural overhead in these models. Model M6 is essentially EfficientLPS with a runtime of $\SI{212.76}{\milli\second}$}.  In the following sections, we further analyze the individual architectural components of the $M5$ model in more detail.

\subsubsection{Influence of Proximity Convolution Parameters}
\label{sec:pil}

\begin{table}
\centering
\caption{Effect of the search area and kernel size on the performance of the proximity convolution module.  All scores are in [\%].} 
\label{tab:PAC ablation}
\begin{tabular}{@{}c|cc|ccc|c@{}}
\toprule
Model & Search grid& Kernel size& PQ  & PQ\textsuperscript{St} & PQ\textsuperscript{Th} & mIoU \\
\midrule
M5$_{1}$ &- & 3x3 & $56.6$& $55.7$& $57.9$ & $61.8$ \\
M5$_2$ &5x5 & 3x3 & \textbf{57.4}& $56.5$ & \textbf{58.7} &  \textbf{62.5} \\
M5$_3$ &7x7 & 3x3 & $57.1$ & \textbf{56.8} & $57.4$ & \textbf{62.5} \\
M5$_4$ &7x7 & 5x5 & \textbf{57.4} & $56.5$ & $58.5$ & $62.2$ \\
\bottomrule
\end{tabular}
\end{table}

The proposed Proximity Convolution (PC) is the core of the proximity convolution module. It employs the kNN algorithm to find the $k$ closest neighbors of each pixel in the projected image where value of $k$ is the product of kernel size dimensions of the proximity convolution. This algorithm is also parameterized by the search grid size that defines a grid around a pixel within which the algorithm performs the search. \tabref{tab:PAC ablation} presents the results of the experiment where we vary the search grid and kernel sizes in the M5 model from \secref{sec:detailedAnalysis}. In the first model M5$_1$, we employ the standard convolution with a kernel size of $3\times3$ in the proximity convolution module of model M5. This model achieves a $PQ$ score of $56.6\%$ and a $mIoU$ of $61.8$. The second model M5$_2$ uses the PC with a search grid of $5\times5$ and a kernel size of $3\times3$. The model M5$_2$ with the proximity convolution achieves an improvement of $0.8\%$ in the $PQ$ score. Here, more than one-third of the search space can be captured by the convolution weights.

In model M5$_3$, we increase the search grid of the PC from $5\times5$ to $7\times7$ while keeping the kernel size the same as model M5$_2$. Here, almost one-fifth of the search space can be captured by the convolution weights. Hence, although the search area increases, there are not enough convolution weights to efficiently capture all the close neighbors. This leads to a decrease in performance of $0.3\%$ in the $PQ$ score. The model M5$_4$ increases the kernel size of the PC in model M5$_3$ to $5\times5$. This model performs similar to the model M5$_2$, although the convolution weights can capture half of the search space. This result indicates that predominantly the top nine neighbors computed in model M5$_2$ were adequate to capture the underlying transformations. Therefore, in the proposed EfficientLPS architecture, we employ the search grid size of $5\times5$ and a proximity convolution kernel size of $3\times3$.
 
\subsubsection{Evaluation of Range Encoder Network}
\label{sec:RFPN}

\begin{table}
\centering
\caption{Influence of the range encoder on the overall panoptic segmentation performance. All scores are in [\%].} 
\label{tab:FPN ablation}
\begin{tabular}{@{}l|ccc|cc|c@{}}\toprule
&  PQ  & PQ\textsuperscript{St}  & PQ\textsuperscript{Th} & SQ & RQ & mIoU \\
\midrule
Downsampled & $54.8$& $54.6$& $55.1$ & $74.1$ & $65.2$ & $59.5$ \\
Range-Encoded & \textbf{57.4}& \textbf{56.5}& \textbf{58.7} & \textbf{75.0} & \textbf{67.6} &\textbf{62.5} \\
\bottomrule
\end{tabular}
\end{table}

The Range Encoder Network (REN) is a small CNN based on the EfficientNet~\cite{tan2019efficientnet} topology which encodes range information. We use compound scaling coefficients of $0.1$ for the width, $0.1$ for the depth, and $224$ for the resolution. Our proposed range-aware FPN and the distance-dependent semantic head both employ the multi-scale features encoded by the REN. The multi-scale features of the REN reinforce coherently aggregated fine and contextual output features of the 2-way FPN with spatial awareness. The dilation offsets for the range-guided depth-wise atrous separable convolutions that are employed at different scales are computed in the distance-dependent semantic head. In this experiment, we show the importance of the REN features compared to the direct incorporation of the range channel in the EfficientLPS architecture. \tabref{tab:FPN ablation} presents the results from this experiment. 

We compare the performance of two models in this experiment. The first model referred to as downsampled, removes the REN from the model M5 and directly downsamples the range channel with downsampling factors $\times4$, $\times8$, $\times16$, and $\times32$. These downsampled versions of the range channels are then fed to the relevant modules. This model achieves a $PQ$ score of $54.8\%$ and a $mIoU$ of $59.5\%$. The second model is essentially the model M5 that already has the REN as part of the network, and we refer to it as the range-encoded model. This model achieves a $PQ$ score of $57.4\%$ and a $mIoU$ of $62.5\%$. The range-encoded model achieves an improvement in $PQ$ and $mIoU$ scores of $2.6\%$ and $3.0\%$ respectively, demonstrating that the direct downsampling is not sufficient to propagate the spatial information to the main network and a dedicated encoder such as the REN is required.

\subsubsection{Evaluation of Range-Aware FPN}
\label{sec:RFPN_fusion}

\begin{table}
\centering
\caption{Evaluation of different methods to incorporate learned range features in the RFPN. All scores are in [\%].} 
\label{tab:fusion ablation}
\begin{tabular}{@{}l|ccc|cc|c@{}}
\toprule
&  PQ  & PQ\textsuperscript{St} & PQ\textsuperscript{Th} & SQ & RQ & mIoU \\
\midrule
Additive & $53.9$& $53.7$& $54.1$ & $74.0$ & $64.6$ & $60.8$ \\
Concatenative & $56.5$& $55.6$& $57.8$ & $74.8$ &  $66.4$&  $61.6$ \\
Fusion & \textbf{57.4} & \textbf{56.5} & \textbf{58.7} & \textbf{75.0} & \textbf{67.6} &\textbf{62.5} \\
\bottomrule
\end{tabular}
\end{table}

There are two main components of our proposed range-aware FPN: the REN and the fusion to incorporate REN features into the FPN. In the previous section, we discuss the importance of the REN for the functioning of range-aware FPN. In this section, we evaluate different fusion methods to incorporate REN features into the FPN. \tabref{tab:fusion ablation} shows the results from this experiment on the M5 model from \secref{sec:detailedAnalysis}. We identify three major techniques to fuse features from multiple network streams: Addition, Concatenation, and Feature fusion. For the additive model, the REN features at each scale are expanded to 256 channels and are summed with the corresponding output of the 2-way FPN to yield the output of the range-aware FPN. In the case of the concatenative model, the outputs of REN and 2-way FPN are concatenated at each scale, followed by two sequential $3\times3$ convolution layers to yield the final output of the range-aware FPN. Each of the convolution layers is followed by an iABNsync and leaky ReLU activation. For the feature fusion model, we employ the mechanism from \eqref{eq:range_fpn}. 

The additive model yields the lowest performance with a $PQ$ score of $53.9\%$ as it treats the features from the REN and the main network equally, even though the representational capacity of the two networks are significantly different. Instead of supporting the main network to capture richer representation, the REN reduces the quality of the overall features. The concatenative model yields a better performance with a $PQ$ score of $56.5\%$. This model provides the network with the flexibility of determining which features are more significant to impart the required spatial awareness and hence achieves a substantial improvement over the additive model. Nevertheless, the fusion model with a $PQ$ score of $57.4\%$ outperforms the other fusion methods. This model further makes the fusion more flexible by extending the concatenative model with additional selectivity control which improves the performance. We expect that more adaptive fusion techniques~\cite{valada2019self,valada2016convoluted,valada2016towards} can further improve the performance of our range-aware FPN.

\subsubsection{Evaluation of Different Semantic Head Topologies}
\label{sec:semantichead}

\begin{table}
\centering
\caption{Evaluation of different variations of the proposed semantic head. All scores are in [\%].} 
\label{tab:RDPC ablation}
\begin{tabular}{@{}cc|ccc|cc|c@{}}
\toprule
RLSFE& RDPC&  PQ  & PQ\textsuperscript{St} & PQ\textsuperscript{Th} & SQ & RQ & mIoU \\
\midrule
\xmark & \xmark & $56.2$& $55.5$& $57.1$ & $74.7$ &  $66.8$ &  $61.1$    \\
\cmark & \xmark & $56.6$& $56.0$ &  $57.4$ & $74.7$ &  $67.0$ & $61.4$   \\
\xmark & \cmark & $57.1$& $56.2$ &  $58.4$ & $74.9$ &  $67.5$ & $62.1$   \\
\cmark & \cmark & \textbf{57.4} & \textbf{56.5} & \textbf{58.7} & \textbf{75.0} & \textbf{67.6} &  \textbf{62.5}   \\
\bottomrule
\end{tabular}
\end{table}

The semantic head incorporates our proposed range-guided variants of the dense prediction cells (DPC)~\cite{chen2018searching} and the large scale feature extractor (LSFE)~\cite{mohan2020efficientps} modules with the bottom-up path connections. In this section, we compare the performance of the original versions of each of these two modules with their proposed range-guided counterparts. \tabref{tab:RDPC ablation} presents the results of this experiments. We train model M5 from \secref{sec:detailedAnalysis} with the semantic head consisting of the original DPC and LSFE modules. This model attain a $PQ$ score of $56.2\%$, $PQ_{st}$ score of $55.5\%$, $PQ_{th}$ score of $57.1\%$, and an $mIoU$ of $61.1\%$. For the second model, we replace the original LSFE module in the first model with our range-guided LSFE (RLSFE) module. This model achieves an improvement of $0.4\%$, $0.5\%$, $0.3\%$, and $0.3\%$ in the $PQ$, $PQ_{st}$, $PQ_{th}$ and $mIoU$ scores respectively, compared to the first model.

We observe a similar improvement in performance of the third model that employs our proposed range-guided DPC module (RDPC). This model achieves an improvement of $0.9\%$ in the PQ score and $1.3\%$ in the $mIoU$ compared to the first model. Interestingly, the improvement in the PQ$^{th}$ score is higher than the improvement in the PQ$^{st}$ score which is $0.7\%$ and $1.3\%$ respectively. This indicates a larger improvement in semantic segmentation of 'thing' classes with the denser and relatively larger receptive field of RDPC compared to 'stuff' classes. Finally, we train the first model with both RLSFE and RDPC modules which achieves a further improvement with a $PQ$ score of $57.4\%$ and an $mIoU$ of $62.5\%$. This experiment demonstrates that our semantic head effectively learns scale-invariant features due to its distance-dependent receptive fields.

\subsubsection{Influence of Panoptic Periphery Loss}

\begin{table}
\centering
\caption{Evaluation of the panoptic periphery loss using the border IoU metric for the person~(Pe), car~(Ca), bicyclist~(Bi) and other vehicle~(Ov) classes. All scores are in [\%].} 
\label{tab:Periphery loss}
\begin{tabular}{c|c|cccc|c@{}}
\toprule
Model & Per. loss& bIoU\textsuperscript{Pe} & bIoU\textsuperscript{Ca} & bIoU\textsuperscript{Bi} & bIoU\textsuperscript{Ov} & PQ\textsuperscript{Th}    \\
\midrule
M4 & \xmark & 66.8  & 89.6 & 74.3 & 40.9 & 56.9 \\
M5 & \cmark & \textbf{70.0}& \textbf{90.3}  & \textbf{75.1}  & \textbf{47.5} &  \textbf{58.7}  \\
\bottomrule
\end{tabular}
\end{table}

We evaluate the boundary refinement performance due to the panoptic periphery loss using the \textit{border-IoU} metric. The \textit{border-IoU} metric enables us to analyze the bleeding or shadowing effects that are observed while projecting the predicted labels back to point clouds. Our proposed loss exploits spatial information to refine the boundaries of panoptic 'thing' class objects. \tabref{tab:Periphery loss} presents the results using the \textit{border-IoU} metric for the top four 'thing' classes that achieve the highest improvement, namely person, car, bicyclist, and other-vehicle.

We compute the \textit{border-IoU} for a border width of 2 pixels for the model M4 and the model M5 from \secref{sec:detailedAnalysis} which are trained with and without the panoptic periphery loss. We observe the highest improvement of $6.6\%$ in $bIoU$ for the other-vehicle class, followed by $3.2\%$ improvement in $bIoU$ for the person class. We also observe an improvement for the car and bicyclist class. The other-vehicle class consists of different types of vehicles such as a trailer, bus, and train, which makes it challenging to accurately segment the boundaries. Similarly, the person class is often only represented with few pixels for extended body parts which again makes it challenging to accurately segment these boundaries. The inaccuracies in the border segmentation are further exacerbated while projecting back to the 3D domain. Hence, explicitly focusing on refining the boundaries using our proposed periphery loss yields substantial improvement.

\subsubsection{Evaluation of Pseudo Labeling}
\label{sec:psuedolabel}

\begin{figure*}
\centering
\footnotesize
{\renewcommand{\arraystretch}{1}
\begin{tabular}{P{0.1cm}P{5.5cm}P{5.5cm}P{5.5cm}}
&  \raisebox{-0.4\height}{Input} &  \raisebox{-0.4\height}{Naive}&  \raisebox{-0.4\height}{Heuristic-Based (ours)} \\
\\
{\rotatebox[origin=c]{90}{Example 1}} &\raisebox{-0.4\height}{\includegraphics[width=\linewidth]{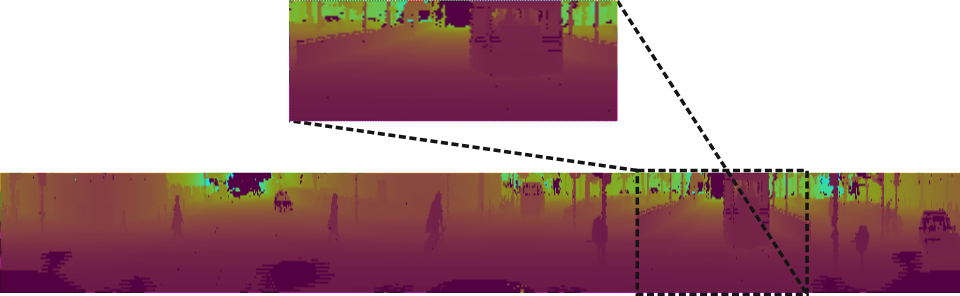}} & \raisebox{-0.4\height}{\includegraphics[width=\linewidth]{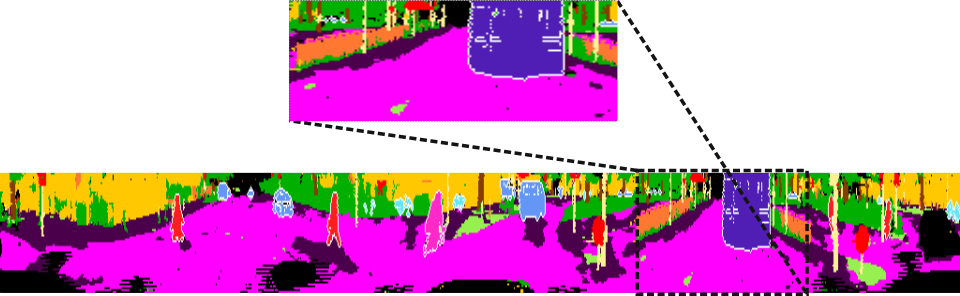}} & \raisebox{-0.4\height}{\includegraphics[width=\linewidth]{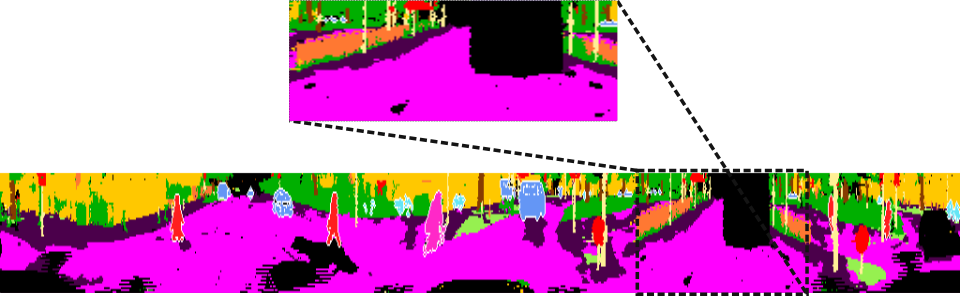}} \\
\\
{\rotatebox[origin=c]{90}{Example 2}} &\raisebox{-0.4\height}{\includegraphics[width=\linewidth]{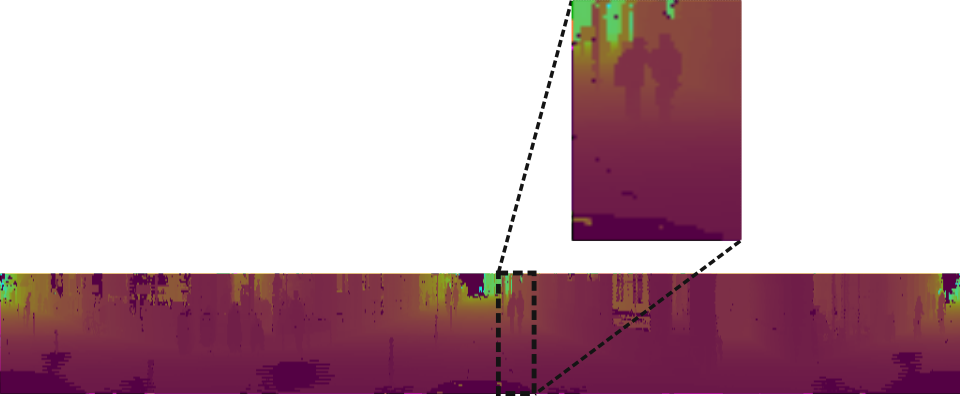}} & \raisebox{-0.4\height}{\includegraphics[width=\linewidth]{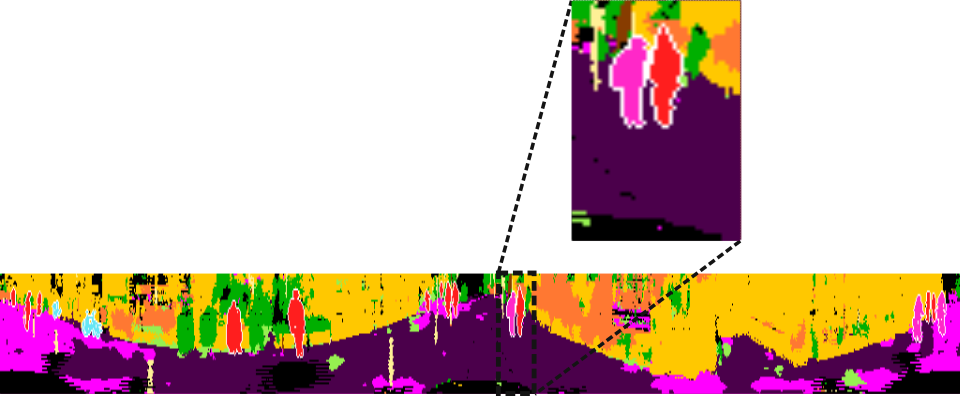}} & \raisebox{-0.4\height}{\includegraphics[width=\linewidth]{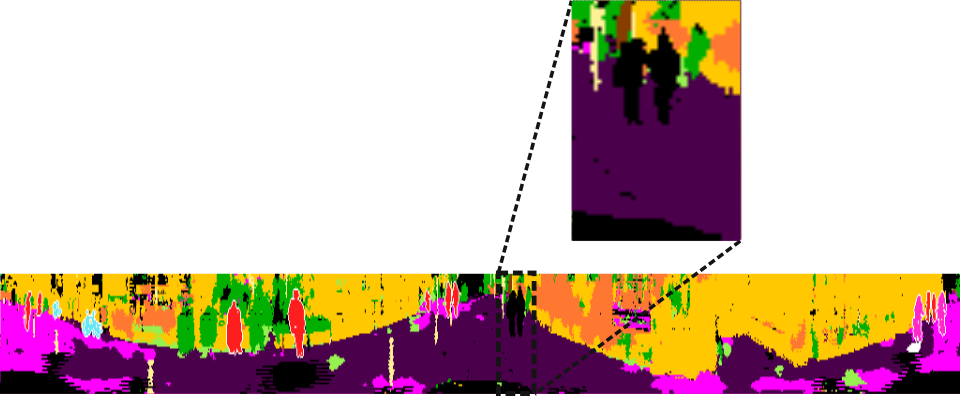}} \\
\end{tabular}}
\caption{Comparison of pseudo labels generated naively and using our proposed heuristic. On the left, the full range image is shown with zoomed in regions. Each example shows the predictions of the full image with zoomed in regions and the input range image. The labels generated using the naive heuristic misclassifies the truck (purple) in example~1 and a person as bicyclist (light purple) in example~2. Our proposed heuristic assigns the misclassified pixels as unlabeled in both examples.}
\label{fig:pseudo_ablation}
\end{figure*}

In this section, we evaluate the performance of our proposed heuristic for generating pseudo labels from unlabeled datasets. We first define two pseudo label generators, one for generating naive pseudo labels (PLG\textsubscript{N}) and another using our proposed heuristic (PLG\textsubscript{O}). PLG\textsubscript{N} is the M5 model from \secref{sec:detailedAnalysis} which achieves the highest $PQ$ score on the SemanticKITTI validation dataset. Whereas, PLG\textsubscript{O} is the M5 models with its hyperparameters set such that it maximizes the ratio $(TP-FP)/TP$ on the validation dataset. The hyperparameters that we tune via grid search are the overlap threshold, minimum stuff area, confidence threshold, and softmax threshold, for the panoptic fusion module; NMS IoU, and score threshold, for the RCNN; the number of proposals for the RPN. For naive labeling of the unlabeled dataset, we use the output of PLG\textsubscript{N} as the final pseudo labels. For our heuristic-based labeling, we employ the post-processing technique described in \secref{sec:psuedolabel} on the predictions of PLG\textsubscript{O} to obtain the final pseudo labels. \figref{fig:pseudo_ablation} shows example pseudo labels generated from both the methods. In example~1, the train is misclassified as a truck in the output from the naive approach, whereas it is classified as unlabeled in the output from heuristic-based method. In example~2, the naive approach misclassifies one of the two persons as a bicyclist, and our heuristic-based approach classifies both the people as unlabeled. These examples demonstrate that our heuristic-based approach assigns objects as unlabeled than risking misclassification wherever possible.

\begin{table}
\centering
\caption{Evaluation of the proposed heuristic for pseudo labeling. All scores are in [\%].} 
\label{tab:heuristic ablation}
\begin{tabular}{@{}c|ccc|cc|c@{}}
\toprule
Heuristic & PQ & PQ\textsuperscript{St} & PQ\textsuperscript{Th} & SQ & RQ & mIoU \\
\midrule
\xmark & $57.4$& $56.5$& $58.7$ & \textbf{75.0}& $67.6$& $62.5$\\
M\textsubscript{naive} & $58.0$& $57.2$& $59.1$ & \textbf{75.0}& $69.3$& $64.0$\\
M\textsubscript{ours} & \textbf{59.2} & \textbf{58.0} & \textbf{60.9} &\textbf{75.0} &\textbf{69.8}& \textbf{64.9} \\
\bottomrule
\end{tabular}
\end{table}

\tabref{tab:heuristic ablation} presents the quantitative results from this experiment. Both M\textsubscript{naive} and M\textsubscript{ours} models have the same architecture as the M5 model from \secref{sec:detailedAnalysis} and are trained using the training scheme described in \secref{sec:psuedolabel}. The difference between the two models is that M\textsubscript{naive} is trained using the naive approach, whereas M\textsubscript{ours} is trained using our heuristic-based technique. We observe that both the models achieve a higher $PQ$ score than the model trained without the pseudo labeled dataset. Moreover, our M\textsubscript{ours} achieves the highest $PQ$ score of $59.2\%$ which is an improvement of $1.8\%$ over M\textsubscript{naive}. These results demonstrate that training our model on a pseudo labeled dataset improves the performance and we obtain a larger improvement if we further optimize by employing a form of regularization on the pseudo labels such as using our proposed heuristic.

\subsubsection{Generalization Ability of Proposed Modules}

\begin{table}
\centering
\caption{Evaluation of the generalization ability of our proposed architectural components. All scores are in [\%].} 
\label{tab:robust ablation}
\begin{tabular}{@{}l|ccc|cc|c@{}}
\toprule
Model & PQ & PQ\textsuperscript{St} & PQ\textsuperscript{Th} & SQ & RQ & mIoU \\
\midrule
Panoptic FPN\textsubscript{vanilla} & $48.7$& $49.6$& $47.5$ & $71.7$& $63.0$& $55.2$ \\
Seamless\textsubscript{vanilla} & $50.6$& $50.7$& $50.5$ & $72.8$& $63.6$& $56.9$ \\
\midrule
Panoptic FPN\textsubscript{ours} & $50.8$& $50.9$& $50.6$ & $72.5$& $63.4$& $56.3$ \\
Seamless\textsubscript{ours} & $53.4$& $52.9$& $54.1$ & $73.1$& $64.2$ & $58.4$ \\
\bottomrule
\end{tabular}
\end{table}

In this experiment, we study the effectiveness and generalization ability of our proposed modules by directly incorporating them into other well-known top-down image panoptic segmentation networks. We choose two state-of-the-art panoptic image segmentation networks: Panoptic~FPN~\cite{kirillov2019panoptice} and Seamless~\cite{porzi2019seamless}. In the vanilla version of both the networks, we use the panoptic fusion module to compute the final panoptic segmentation predictions. We train the vanilla versions with an input resolution of $4096\times256$ using scan unfolding based projection and kNN-based post-processing. We employ Lov\'{a}sz Softmax in addition to weighted per-pixel log loss during training. \tabref{tab:robust ablation} shows the results from this experiment. The Panoptic~FPN\textsubscript{vanilla} model achieves a $PQ$ score of $48.7\%$ and an $mIoU$ of $55.2\%$. While the Seamless\textsubscript{vanilla} model achieves a $PQ$ score of $50.6\%$ and an $mIoU$ of $56.9\%$.

\begin{figure*}
\centering
\footnotesize
\setlength{\tabcolsep}{0.1cm}
{\renewcommand{\arraystretch}{0.5}
\begin{tabular}{P{0.4cm}P{5.5cm}P{5.5cm}P{5.5cm}}
& \raisebox{-0.4\height}{Baseline Output} & \raisebox{-0.4\height}{EfficientLPS Output (Ours)} &  \raisebox{-0.4\height}{Improvement/Error Map} \\
\\
{\rotatebox[origin=c]{90}{Semantic KITTI (a)}} &\raisebox{-0.4\height}{\includegraphics[width=\linewidth]{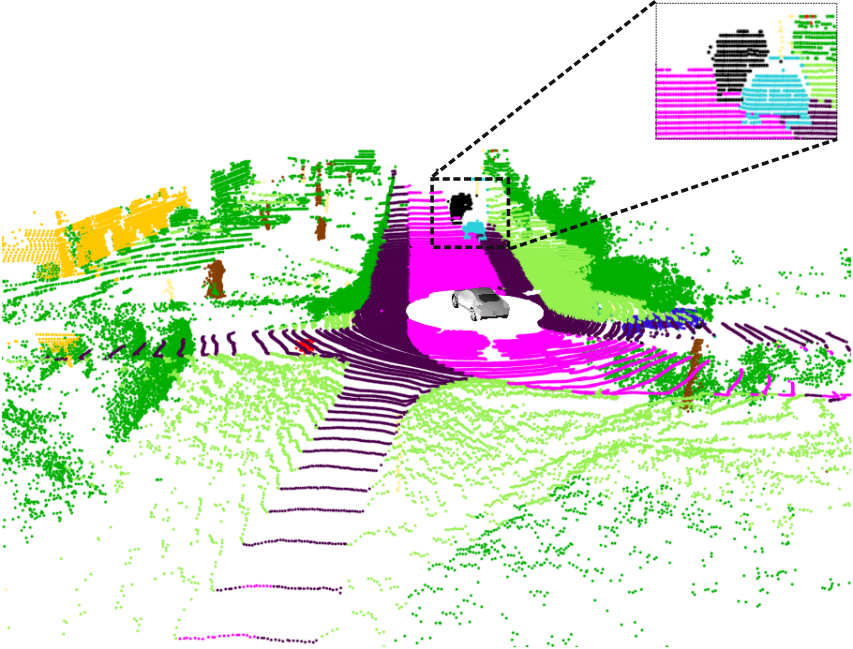}} & \raisebox{-0.4\height}{\includegraphics[width=\linewidth]{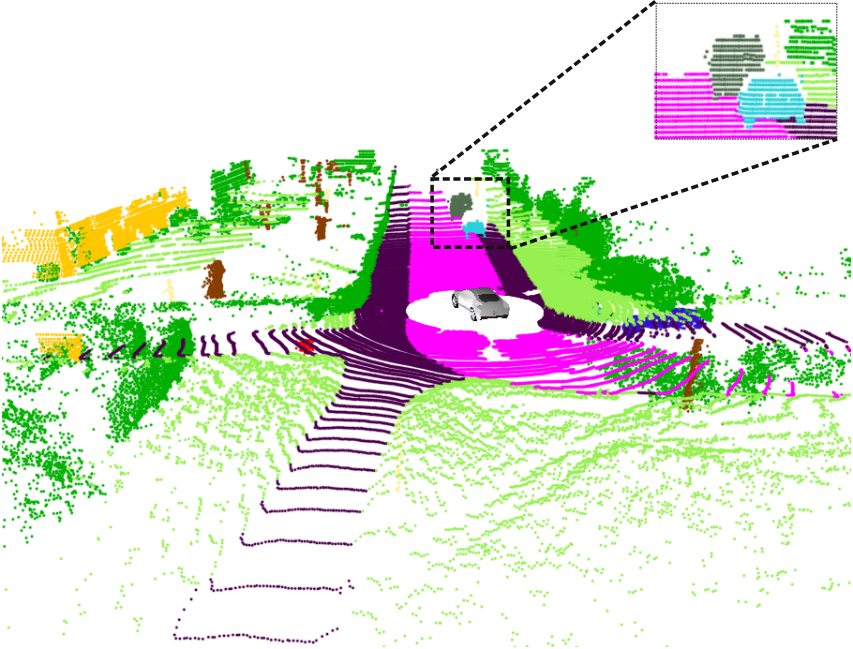}} & \raisebox{-0.4\height}{\includegraphics[width=\linewidth]{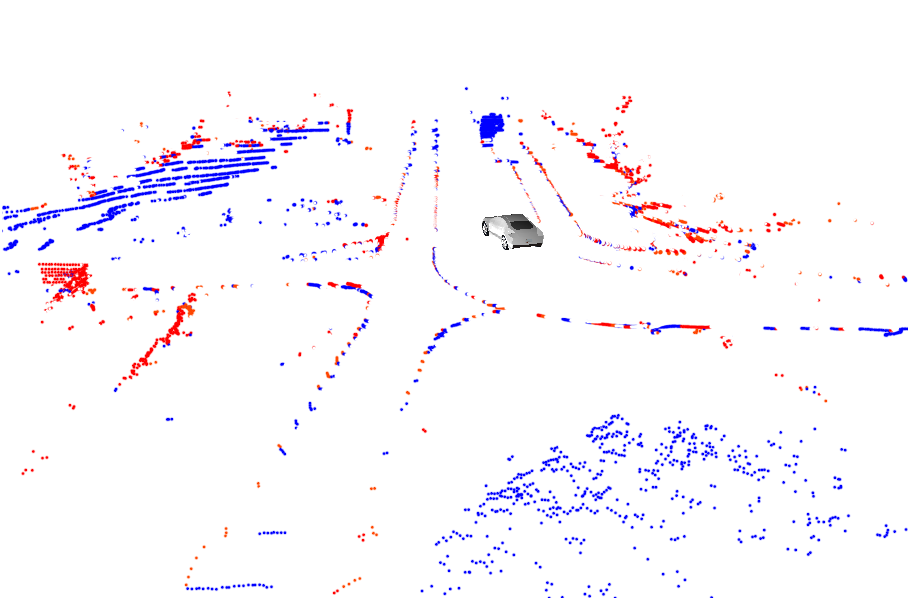}} \\
\\
{\rotatebox[origin=c]{90}{Semantic KITTI (b)}}&\raisebox{-0.4\height}{\includegraphics[width=\linewidth]{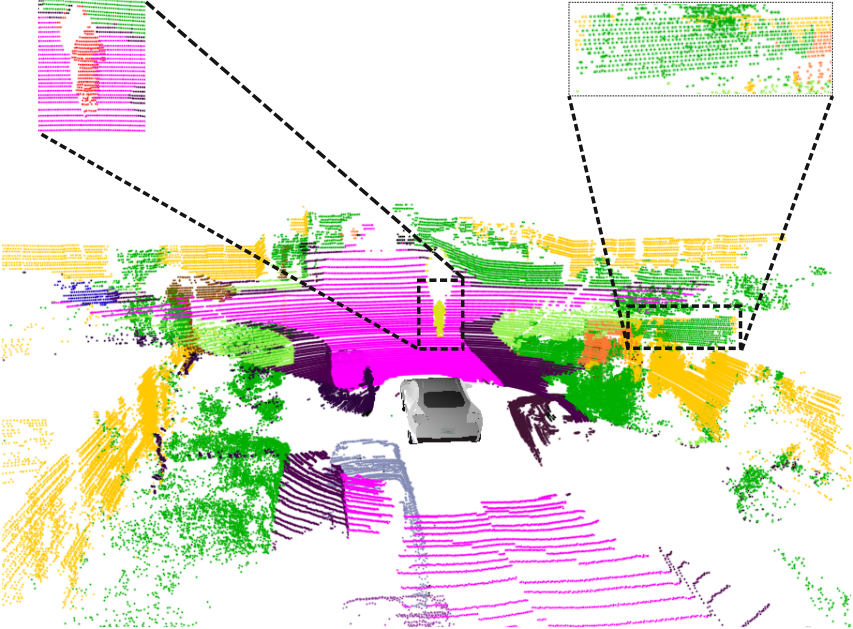}} & \raisebox{-0.4\height}{\includegraphics[width=\linewidth]{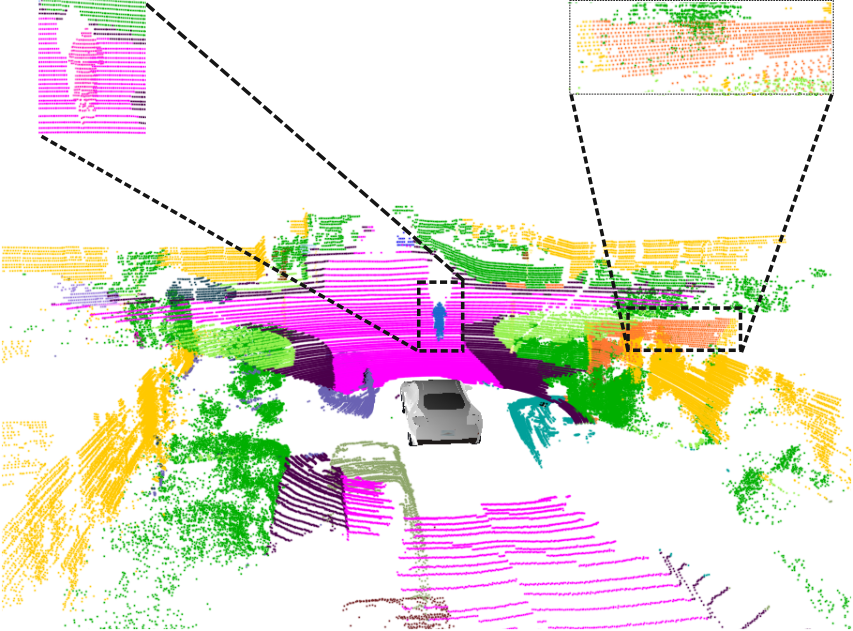}} & \raisebox{-0.4\height}{\includegraphics[width=\linewidth]{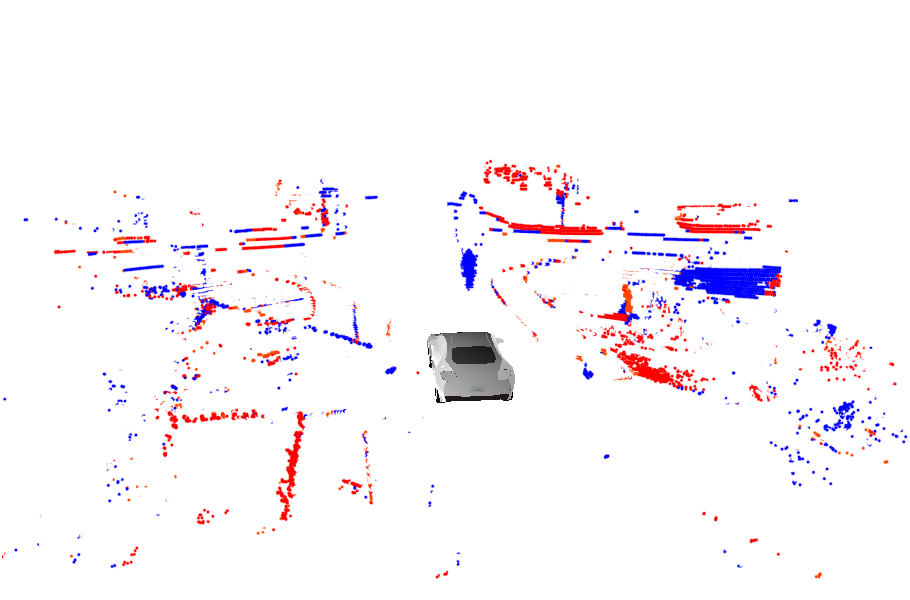}} \\
\\
{\rotatebox[origin=c]{90}{nuScenes (c)}} &\raisebox{-0.4\height}{\includegraphics[width=\linewidth]{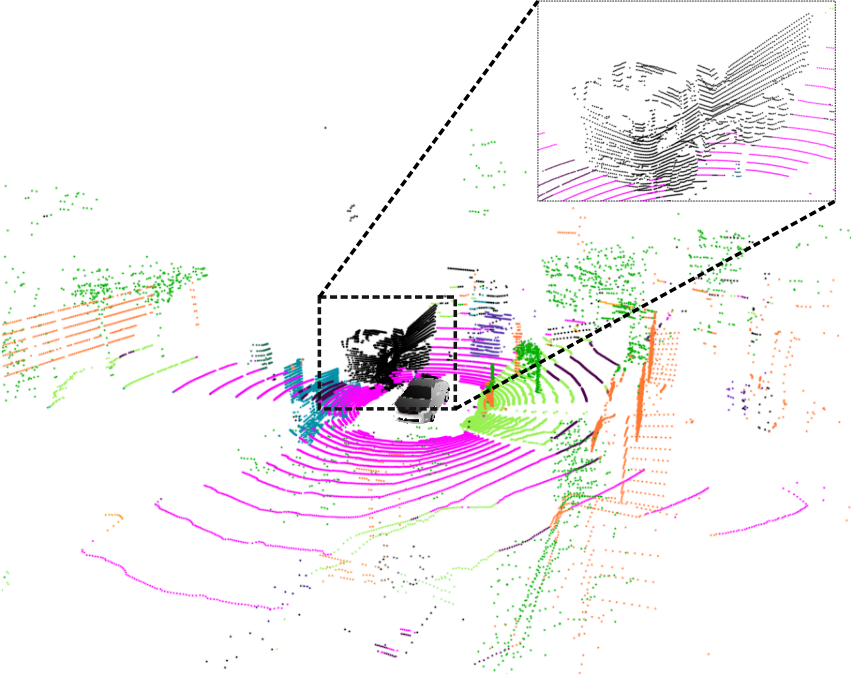}} & \raisebox{-0.4\height}{\includegraphics[width=\linewidth]{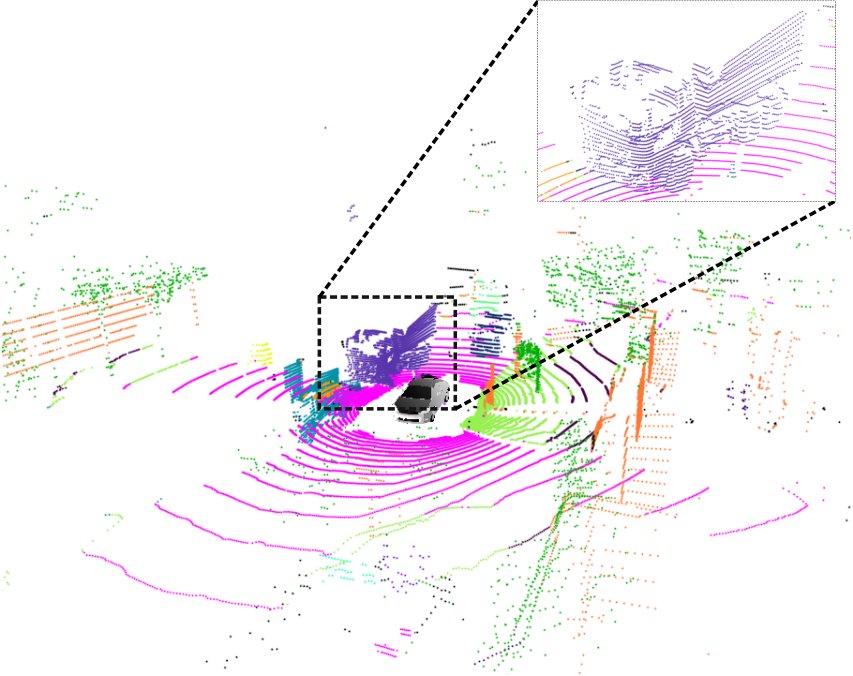}} & \raisebox{-0.4\height}{\includegraphics[width=\linewidth]{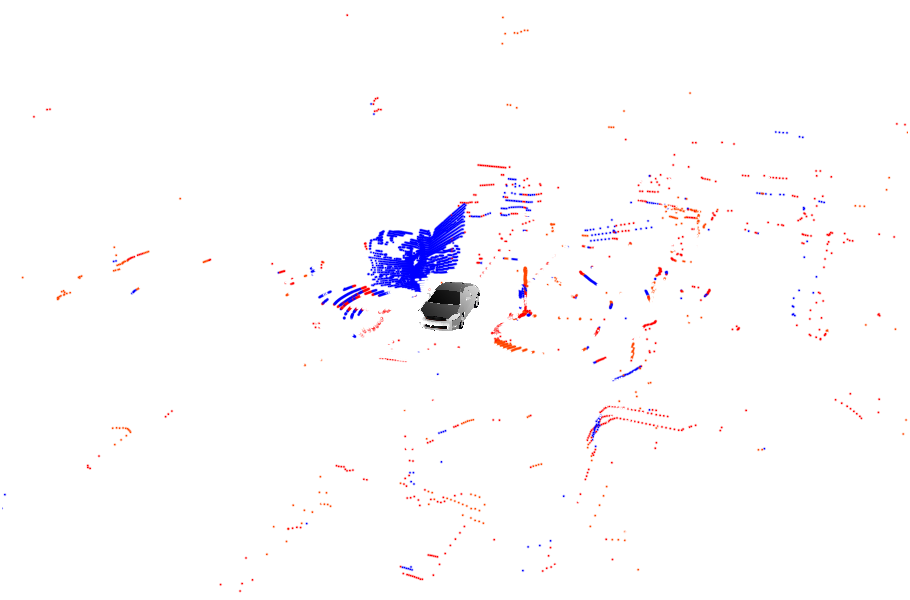}} \\
\\
{\rotatebox[origin=c]{90}{nuScenes (d)}} &\raisebox{-0.4\height}{\includegraphics[width=\linewidth]{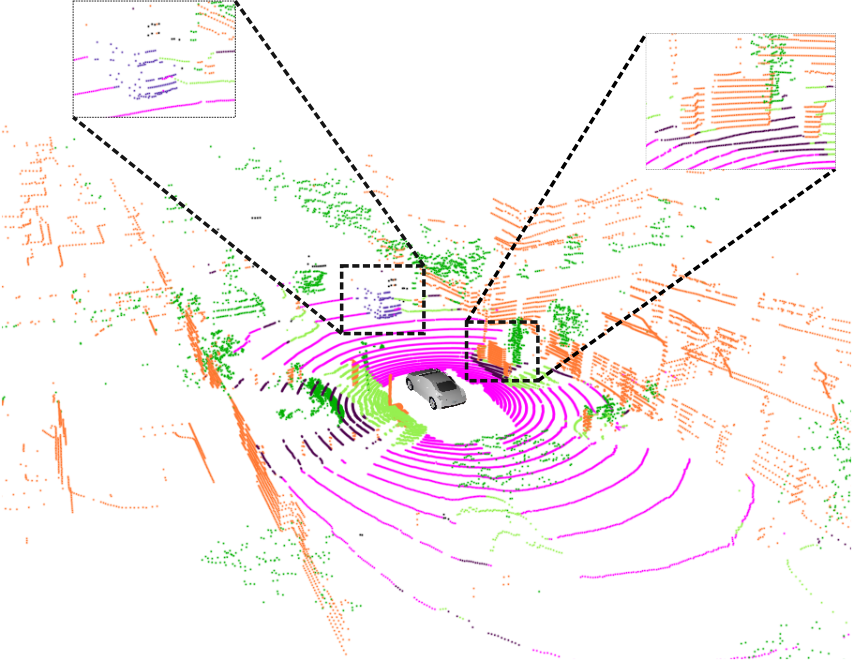}} & \raisebox{-0.4\height}{\includegraphics[width=\linewidth]{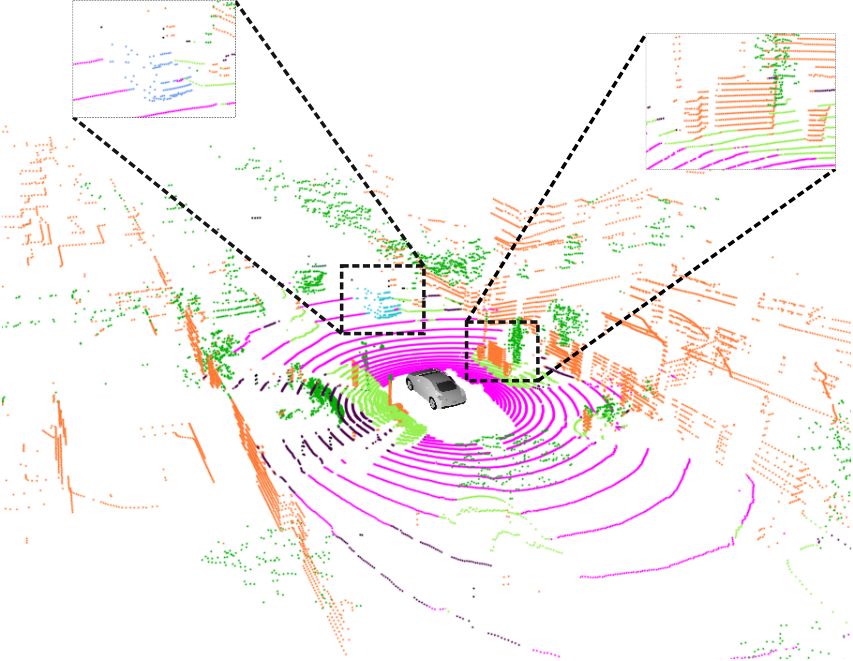}} & \raisebox{-0.4\height}{\includegraphics[width=\linewidth]{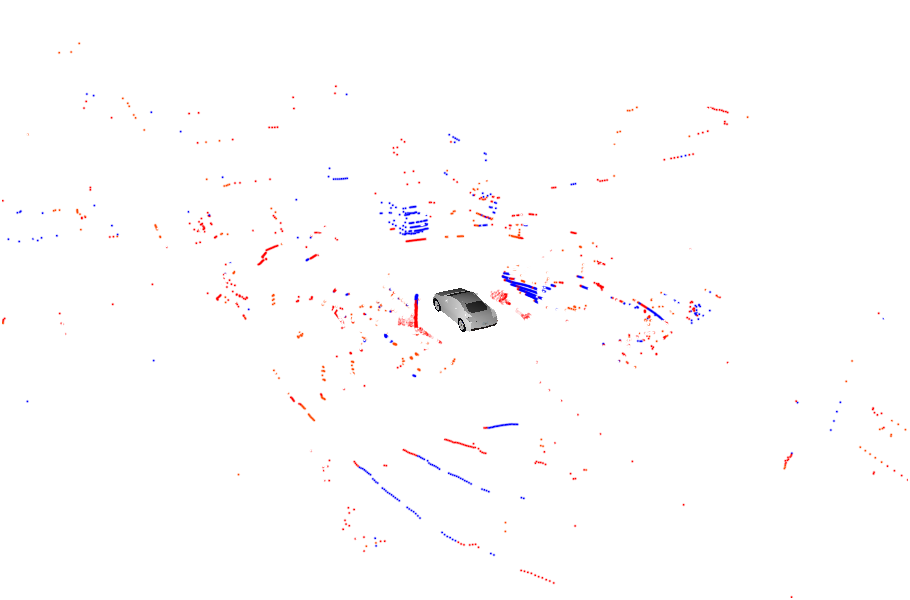}} \\
\end{tabular}}
\caption{Qualitative comparison of LiDAR panoptic segmentation results. We compare against PanopticTrackNet on SemanticKITTI and \mbox{(KPConv + Mask~R-CNN)} on nuScenes validation sets. We also show the improvement/error map which shows the points that are misclassified by EfficientLPS in red and the points that are misclassified by the baseline but correctly predicted by EfficientLPS in blue.}
\label{fig:semantic_qualitative}
\end{figure*}

In our version of both the networks, we introduce the proximity convolution module before the encoder, followed by employing the REN in parallel to the main encoder and switching the standard FPN to our range-aware FPN. Additionally, we use the panoptic periphery loss function during training. Since Panoptic~FPN only employs a $3\times3$ convolution in the semantic head for each of its FPN scales, we replace this convolution with our proposed $3\times3$ range-guided depth-wise atrous separable convolution with $D_{max}=3$. We keep $D_{max}$ value low to compute denser features with a small receptive field in contrast to a high value that will output sparse features with a large receptive field. In the case of the Seamless network, we extend the miniDL module in the semantic head with an additional parallel branch that consists of our proposed $3\times3$ range-guided depth-wise atrous separable convolution with $D_{max}=12$. As a result, Panoptic~FPN\textsubscript{ours} achieves an improvement of $2.1\%$ in the $PQ$ score and $1.1\%$ in $mIoU$ over the vanilla version. We also observe a similar improvement of $2.8\%$ in the $PQ$ score and $1.5\%$ in $mIoU$ for the Seamless\textsubscript{ours} model. This demonstrates the generalization ability of our proposed architectural modules as the direct incorporation of these modules without any tuning of parameters still achieves substantial improvement over the vanilla version. These results also show that any future top-down panoptic segmentation network can easily be transformed into a LiDAR panoptic segmentation network by incorporating our proposed architectural components.

\subsection{Qualitative Evaluations}
\label{sec:qualitative}

In this section, we qualitatively evaluate the panoptic segmentation performance of our proposed EfficientLPS model on the validation set of SemanticKITTI and nuScenes datasets. We compare the results of EfficientLPS with the best performing baseline from \secref{sec:comparisonSOTA} for the respective datasets. To this end, we compare with PanopticTrackNet and (KPConv + Mask~R-CNN) for SemanticKITTI and nuScenes datasets respectively. \figref{fig:semantic_qualitative} shows two comparisons for each of these datasets. We also present the improvement/error map for each of the comparisons and enlarge the segments of the outputs that show significant misclassification. The improvement/error map depicts the points that are misclassified by EfficientLPS with respect to the groundtruth in red and the points that are correctly predicted by EfficientPS but are misclassified by the baseline model in blue.\looseness=-1

\figref{fig:semantic_qualitative}~(a) and \figref{fig:semantic_qualitative}~(b) show examples from the SemanticKITTI dataset in which the improvement over the baseline output (PanopticTrackNet) can be seen in the more accurate segmentation of the other-vehicle class as well as the improvement in distinguishing inconspicuous 'thing' classes such as person and bicyclist. EfficientLPS also demonstrates a more clear separation of object instances while segmenting cluttered classes. This can be primarily attributed to the proximity convolution module and the range-aware FPN that enables the network to have enhanced transform modeling capacity along with distance-aware semantically rich multi-scale features. This enables our model to learn highly discriminative features to accurately classify semantically related classes but at the same time the spatial awareness allows it to accurately segment object instances that are very close to each other. In \figref{fig:semantic_qualitative}~(a) the baseline fails to classify the other-vehicle class, whereas our model accurately classifies this object. In \figref{fig:semantic_qualitative}~(b), the baseline model segments the bicyclist but classifies is it as a person depicted in red (label color for person class). In contrast, our model correctly classifies it as a bicyclist depicted in magenta (label color for bicyclist class). Our model also successfully segments both vegetation and fence classes, as opposed to the baseline which misclassifies most of the fence as vegetation. Additionally, the effects of the boundary refinement is prominent in the segmentation of the car in \figref{fig:semantic_qualitative}~(a).

In \figref{fig:semantic_qualitative}~(c) and \figref{fig:semantic_qualitative}~(d), we qualitatively compare the performance on the sparse nuScenes dataset. We observe that in \figref{fig:semantic_qualitative}~(c) the truck is not segmented in the output of the baseline model (KPConv + Mask~R-CNN), while our EfficientLPS model accurately segments the instance of the truck. In \figref{fig:semantic_qualitative}~(d) the baseline model segments the oncoming car instance but misclassifies it as a truck. It is significantly hard to accurately classify this instance as only a few points lie on the object. Nevertheless, our model still classifies these points as a car, which can be attributed to the shared backbone and the adaptable dense receptive field of the semantic head as well as the instance head. In \figref{fig:semantic_qualitative}~(d) the terrain class is accurately segmented by the EfficientLPS model, whereas the baseline misclassifies it as sidewalk.

\section{Conclusion}
\label{sec:conclusion}

In this work, we presented a novel top-down approach for LiDAR panoptic segmentation using a 2D CNN that effectively leverages the unique spatial information provided by LiDAR point clouds. Our EfficientLPS architecture achieves state-of-the-art performance by incorporating novel architectural components that mitigate the problems caused by projecting the point cloud into the 2D domain. We proposed the proximity convolution module that effectively models geometric transformations of points in the projected image by exploiting the proximity between points. Our novel range-aware FPN demonstrates effective fusion of range encoded features with that of the main encoder to aggregate semantically rich multi-scale range-aware features. We proposed a new distance-dependent semantic head that incorporates our range-guided depth-wise atrous separable convolutions with adaptive dilation rates that cover large receptive fields densely to better capture the contextual semantic information. We further introduced the panoptic periphery loss function that refines object boundaries by utilizing the range information for maximizing the gap between the object boundary and the background. Moreover, we presented a new heuristic for generating pseudo labels from an unlabeled datasets to assist the network with additional training data.

We introduced panoptic ground truth annotations for the sparse LiDAR point clouds in the nuScenes dataset which we made publicly available. Additionally, we provided several baselines for LiDAR panoptic segmentation on this new dataset. We presented comprehensive benchmarking results on SemanticKITTI and nuScenes datasets that show that EfficientLPS sets the new state-of-the-art. EffcientLPS is ranked~\#1 on the SemanticKITTI panoptic segmentation leaderboard. We presented exhaustive ablation studies with quantitative and qualitative results that demonstrate the novelty of the proposed architectural components. Furthermore, we made the code and models publicly available.

\section*{Acknowledgements}

This work was partly funded by the European Union’s Horizon 2020 research and innovation program under grant agreement No 871449-OpenDR, a research grant from Eva Mayr-Stihl Stiftung, and partly financed by the Baden-Württemberg Stiftung gGmbH.\looseness=-1

\ifCLASSOPTIONcaptionsoff
  \newpage
\fi


\bibliographystyle{IEEEtran}
\bibliography{references.bib}

\begin{IEEEbiography}[{\includegraphics[width=1in,height=1.25in,clip,keepaspectratio]{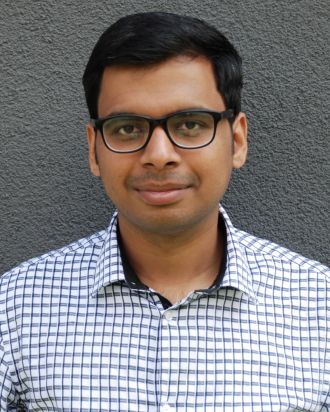}}]{Kshitij Sirohi} is a Ph.D.~student in the Autonomous Intelligent Systems group headed by Wolfram Burgard. He received his M.S.~degree in Embedded Systems Engineering from the University of Freiburg in 2020. His research focuses on Deep learning and probabilistic approaches for robot perception, localization and mapping.
\end{IEEEbiography}

\begin{IEEEbiography}[{\includegraphics[width=1in,height=1.25in,clip,keepaspectratio]{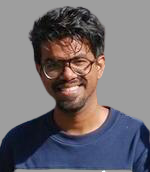}}]{Rohit Mohan} is a Ph.D.~student in the Robot Learning Lab headed by Abhinav Valada. He received his M.S.~degree in Computer Science from the University of Freiburg in 2021. His research focuses on robot perception and navigation using deep learning.
\end{IEEEbiography}

\begin{IEEEbiography}[{\includegraphics[width=1in,height=1.25in,clip,keepaspectratio]{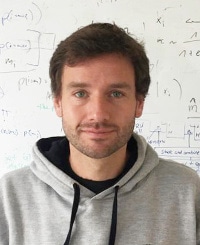}}]{Daniel B\"uscher} is a postdoctoral researcher in the Autonomous Intelligent Systems group headed by Wolfram Burgard. He received his Ph.D.~degree from the University of Freiburg in 2012. His research focuses on autonomous robot navigation and perception using deep learning.
\end{IEEEbiography}

\begin{IEEEbiography}[{\includegraphics[width=1in,height=1.25in,clip,keepaspectratio]{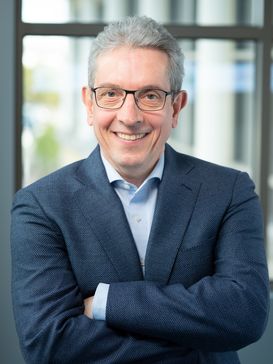}}]{Wolfram Burgard} is Vice President for Automated Driving Technology at the Toyota Research Institute in Los Altos, USA. He is on leave from a Professorship for Computer Science at the University of Freiburg, Germany where he heads the Laboratory for Autonomous Intelligent Systems. He received his Ph.D.~degree in computer science from the University of Bonn in 1991. His areas of interest lie in robotics and artificial intelligence. Wolfram Burgard and his group developed several innovative techniques for robot navigation, perception and control. Wolfram Burgard is head of the BrainLinks-BrainTools Center at the University of Freiburg. He is fellow of the AAAI, the EurAi, the IEEE and the ELLIS Society. Wolfram Burgard is also member of the German Academy of Sciences Leopoldina and the Heidelberg Academy of Sciences and Humanities. 
\end{IEEEbiography}

\begin{IEEEbiography}[{\includegraphics[width=1in,height=1.25in,clip,keepaspectratio]{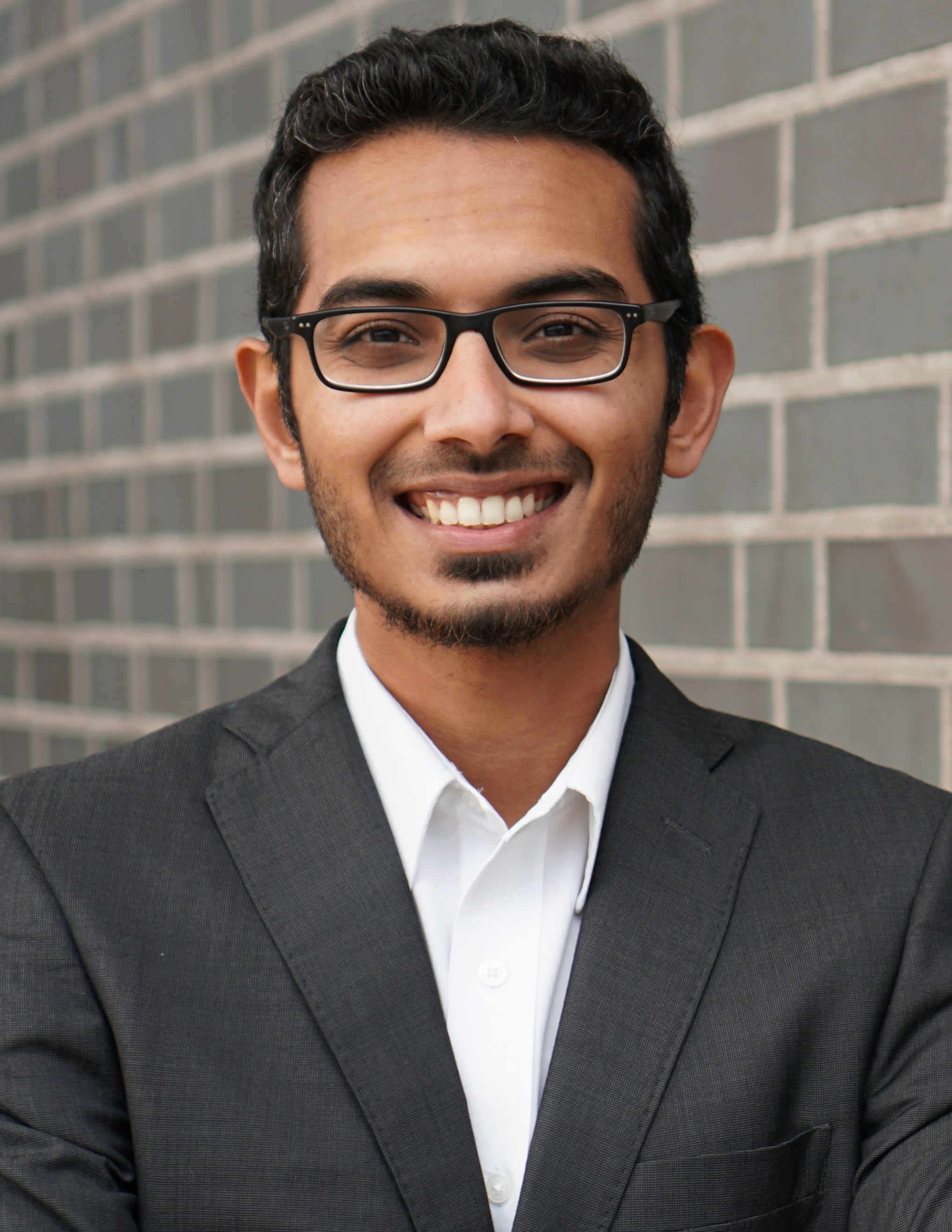}}]{Abhinav Valada}
is an Assistant Professor and head of the Robot Learning Lab at the University of Freiburg. He is a member of the Department of Computer Science, a principal investigator at the BrainLinks-BrainTools Center, and a core faculty in the European Laboratory for Learning and Intelligent Systems (ELLIS) unit in Freiburg. He received his Ph.D.~in Computer Science from the University of Freiburg in 2019 and his M.S.~degree in Robotics from Carnegie Mellon University in 2013. His research lies at the intersection of robotics, machine learning and computer vision with a focus on tackling fundamental robot perception, state estimation and control problems using learning approaches in order to enable robots to reliably operate in complex and diverse domains. Abhinav Valada is a scholar of the ELLIS Society.
\end{IEEEbiography}


\vfill


\end{document}